\documentclass[lettersize,journal]{IEEEtran}
\usepackage[table]{xcolor} 
\usepackage{amsmath,amsfonts}
\usepackage{algorithmic}
\usepackage{algorithm}
\usepackage{array}
\usepackage[caption=false,font=normalsize,labelfont=sf,textfont=sf]{subfig}
\usepackage{textcomp}
\usepackage{stfloats}
\usepackage{booktabs}
\usepackage{url}
\usepackage{verbatim}
\usepackage{graphicx}
\usepackage{cite}
\usepackage{xcolor}
\usepackage{diagbox}
\usepackage{multirow}
\usepackage{pdfpages}

\usepackage[colorlinks=true, linkcolor=blue, citecolor=blue, urlcolor=blue]{hyperref}

\begin{document}

\title{A Feature-level Bias Evaluation Framework for Facial Expression Recognition Models}

\author{ Tangzheng Lian and Oya Celiktutan \\ Centre for Robotics Research, Department of Engineering \\ King’s College London, WC2R 2LS London, U.K. \\ E-mail:\{lian.tangzheng, oya.celiktutan\}@kcl.ac.uk
}

\maketitle



\begin{abstract}
Recent studies on fairness have shown that Facial Expression Recognition (FER) models exhibit biases toward certain visually perceived demographic groups. However, the limited availability of human-annotated demographic labels in public FER datasets has constrained the scope of such bias analysis. To overcome this limitation, some prior works have resorted to pseudo-demographic labels, which may distort bias evaluation results. Alternatively, in this paper, we propose a feature-level bias evaluation framework for evaluating demographic biases in FER models under the setting where demographic labels are unavailable in the test set. Extensive experiments demonstrate that our method more effectively evaluates demographic biases compared to existing approaches that rely on pseudo-demographic labels. Furthermore, we observe that many existing studies do not include statistical testing in their bias evaluations, raising concerns
that some reported biases may not be statistically significant
but rather due to randomness. To address this issue, we introduce a plug-and-play statistical module to ensure the statistical significance of biased evaluation results. A comprehensive bias analysis based on the proposed module is then conducted across three sensitive attributes (age, gender, and race), seven facial expressions, and multiple network architectures on a large-scale dataset, revealing the prominent demographic biases in FER and providing insights on selecting a fairer network architecture.


\end{abstract}


\section{Introduction}

The advancement of deep learning models relies on obtaining extensive datasets in the wild, which are usually gathered through crowd-sourcing or web crawling initiatives \cite{deng2009imagenet}. However, despite its efficacy in gathering a broad spectrum of data, it unintentionally introduces demographic biases\footnote{By demographic biases, we refer to two aspects: at the dataset level, they refer to imbalances in demographic distributions; at the model level, they refer to disparities in performance across different demographic groups.} in the dataset \cite{yang2020towards}. Without explicit intervention, deep learning models trained on biased datasets have been shown to inherit \cite{sagawa2019distributionally} or even amplify \cite{wang2021directional} the biases present in the training data. Consequently, these models exhibit unequal performance across different subgroups of data \cite{eyuboglu2022domino, sohoni2020no}. This issue is particularly concerning when the dataset contains faces, as they inherently include sensitive information. Disparities in the performance of models trained on such face data can lead to discriminatory outcomes against certain perceived\footnote{The demographic groups in this paper are based on visual perception from facial appearances rather than biological sex, self-identified gender, race, or age of the subject. Please see Section~\ref{VI(A)} for a detailed discussion.} demographic groups, thereby undermining fairness, a fundamental principle in the implementation of Trustworthy AI \cite{thiebes2021trustworthy}, Ethical AI \cite{jobin2019global}, or Responsible AI \cite{arrieta2020explainable} systems.

\begin{figure}[tb]
  \centering
    \includegraphics[height=5.5cm, width=8.5cm]{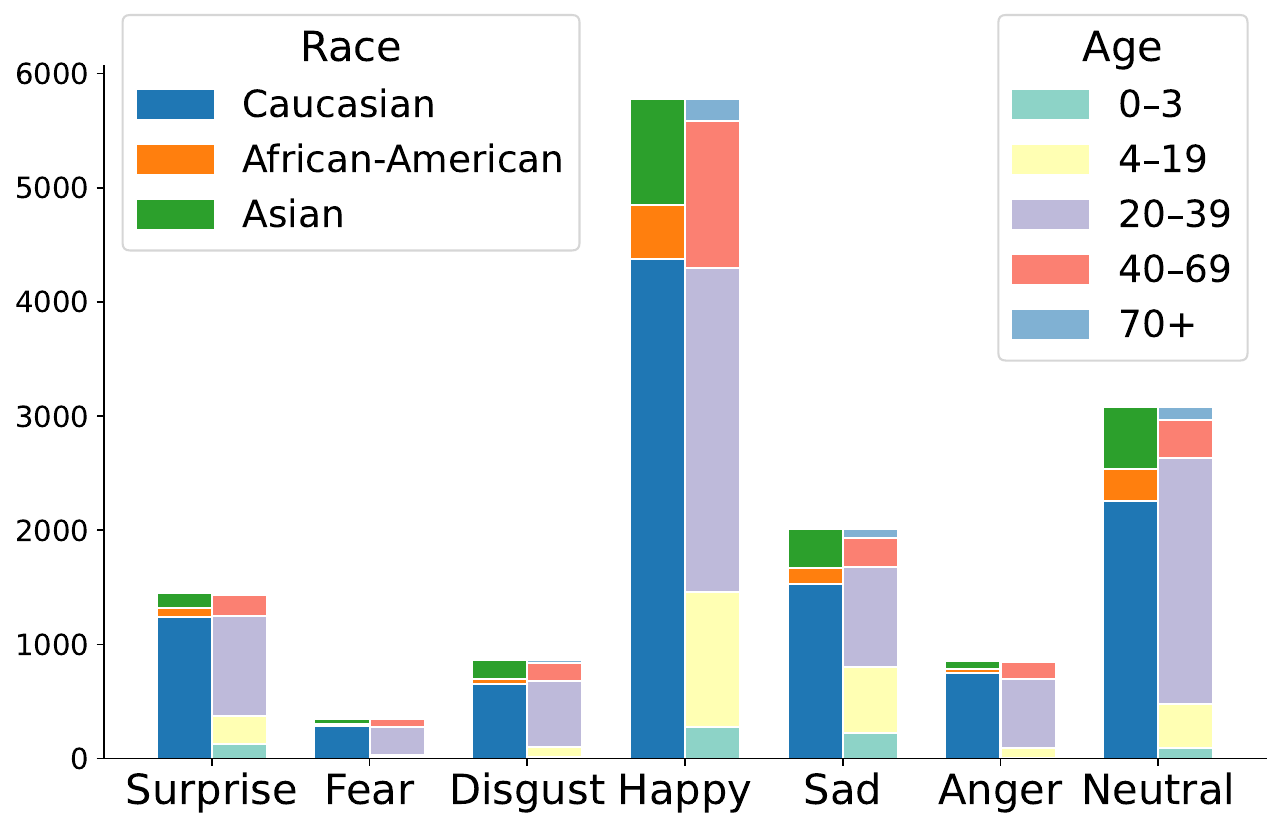}
    
    \caption{Race (left column) and age (right column) distribution for each facial expression in the RAF-DB \cite{li2017reliable} dataset. African-American faces showing fear and very young (0–3 years) or older (70+) faces expressing fear or anger are severely underrepresented, each with fewer than 20 samples. These categories still need to be further split into train/val/test sets.}
    \label{fig:1}
\end{figure}

Recent studies have highlighted disparities in model performance across perceived age, gender, and racial groups in various facial image analysis tasks \cite{singh2022anatomizing}, including facial attribute classification \cite{lin2022fairgrape, park2022fair, ramaswamy2021fair}, face detection \cite{menezes2021bias, yang2022enhancing}, and facial recognition \cite{buolamwini2018gender, wang2020mitigating, wang2019racial, gong2021mitigating, xu2021consistent}. Facial expression recognition (FER) is an essential facial image analysis task to recognize emotional signals from faces and has broad applications in augmented and virtual reality (AR/VR) \cite{barrios2021farapy, ciftci2017partially}, human-computer interaction (HCI) \cite{bartlett2003real, cowie2001emotion}, and interactive therapy systems \cite{ahmed2017automated, barrios2021farapy}. Having a reliable and effective method for evaluating demographic biases in FER models is essential for the responsible development of these models and plays a critical role in preventing such biases from propagating into downstream applications. Previous fairness research in FER \cite{kolahdouzi2023toward, xu2020investigating, cheong2022counterfactual, churamani2022domain, suresh2022using} often relies on the availability of demographic labels in the test set to calculate performance disparities for bias evaluation (Please see Section~\ref{III(A)} for a detailed discussion). However, such demographic labels are typically not available in public FER datasets.

This limitation has narrowed the scope of prior research on fairness in FER. For instance, some studies have limited their focus to a single sensitive attribute \cite{park2022facial, domnich2021responsible, manresa2022facial, dominguez2022gender, dominguez2024less, sham2023ethical}, while others analyze only a subset of facial expressions \cite{fan2021demographic, chen2021understanding}. Even studies that consider all seven basic facial expressions and three sensitive attributes \cite{xu2020investigating, kolahdouzi2023toward, churamani2022domain, cheong2022counterfactual, cheong2023causal} often restrict their experiments to the RAF-DB dataset \cite{li2017reliable}. However, RAF-DB is relatively small, and as shown in Fig.\ref{fig:1}, some categories contain too few samples to draw statistically significant conclusions or capture the full diversity of a visual concept. As a result, bias evaluations in these underrepresented categories may be unreliable due to small sample sizes. Larger FER datasets, such as AffectNet \cite{mollahosseini2017affectnet}, need to be annotated with perceived demographic labels to enable more robust and reliable bias evaluations \cite{hu2025rethinking}. However, annotating perceived demographic labels for large and diverse FER datasets requires significant human effort. Therefore, it is important to develop methods that can still effectively evaluate demographic biases in FER models in cases where the test set does not include demographic labels.

Some studies have attempted to address this issue by training facial attribute classifiers to assign pseudo-demographic labels to images in the test set and then evaluating bias based on these labels \cite{chen2021understanding, dominguez2022gender, dominguez2024metrics, dominguez2024less}.  However, the quality of these pseudo-demographic labels has not been thoroughly evaluated, and they may distort the results of bias evaluation. Additionally, prior studies \cite{xu2020investigating,cheong2022counterfactual,churamani2022domain,suresh2022using,hu2025rethinking} have not applied statistical tests for bias evaluation, raising concerns that some reported biases may not be statistically significant but rather due to randomness. Unlike previous methods that rely on pseudo-demographic labels, we evaluate biases in the feature space of FER models. This idea is inspired by recent advancements in fair representation learning (FRL) \cite{ramaswamy2021fair, quadrianto2019discovering}, which suggest that biases in deep learning models primarily originate from learned representations in the feature space rather than the final classification layer. Bias evaluation in the feature space has been explored in computer vision research \cite{sirotkin2022study, brinkmann2023multidimensional} through the Image Embedding Association Test (iEAT) \cite{steed2021image}. However, iEAT is constrained by a fixed set of target-attribute pairs and is designed exclusively for self-supervised models, making it challenging to apply to FER. 

To address these limitations, we propose a feature-level bias evaluation framework to evaluate the demographic biases in FER under the setting where demographic labels are missing in the test set. Our method evaluates biases by using a probe dataset to compute differential associations in the feature space of FER models, through a reformulation of the iEAT. In addition, we introduce a statistical module to ensure the statistical significance of the reported biases in FER, which can be applied to both existing evaluation pipelines and our proposed framework that analyzes the feature
space. A comprehensive bias analysis is then conducted based on the proposed module. The contributions of this paper can be summarized as follows: 

1. We propose a feature-level
bias evaluation framework to evaluate the demographic biases
in FER in scenarios where demographic labels are not available in the test set.

2. Extensive experiments show
that our method can more effectively evaluate demographic biases in
FER models compared to existing methods using pseudo-demographic labels.

3. We introduce a plug-and-play statistical module to ensure statistically significant bias evaluation results, which can be applied to both existing evaluation pipelines and our proposed framework that analyzes the feature space.

4.  A comprehensive bias analysis based on the statistical module is then conducted across three sensitive attributes (age, gender, and race), seven facial expressions, and multiple network architectures on a large-scale dataset, revealing the prominent demographic biases in FER and providing insights on selecting a fairer network architecture.

\section{Related Work}

\subsection{Demographic Biases in FER}
\subsubsection{Biases in FER Datasets}
Biases in FER datasets can emerge from two primary sources. The first source is data-level bias, which arises during the collection of facial images with various expressions. These images are often obtained through web search engines or web crawling \cite{mollahosseini2017affectnet}. However, web search engines often reflect gender, age, or racial biases \cite{yang2020fairness}, resulting in uneven perceived demographic distributions in the facial images retrieved for each facial expression query. The second source is annotation-level bias, which arises during the labeling process, where annotators assign facial expressions to collected images. Psychological studies have shown that implicit biases, shaped by societal and perceptual factors, significantly influence the interpretation of facial expressions across demographic groups, including age \cite{folster2014facial}, gender \cite{steephen2018we}, and race \cite{hugenberg2004ambiguity}. For example, research indicates that women are often perceived as happier than men \cite{steephen2018we}, and angry expressions are more accurately detected on Black faces, while happy expressions are more easily recognized on White faces \cite{hugenberg2004ambiguity}. These perception biases distort individual annotations and embed systemic biases into the datasets \cite{chen2021understanding}.

\subsubsection{Evaluating and Mitigating Biases in FER Models}
Deep learning models trained on biased datasets often inherit \cite{sagawa2019distributionally} or even amplify \cite{wang2021directional} biases present in the training data. This results in unequal model performance across different subsets of data \cite{eyuboglu2022domino, sohoni2020no}. For instance, if male faces dominate the training data for anger expression, the model may learn to associate anger predominantly with male faces. Consequently, during testing, the model will likely perform better for male faces when predicting anger. Fairness studies in FER can broadly be categorized into two groups: those that aim to evaluate bias \cite{dominguez2022gender, dominguez2024metrics, fan2021demographic, manresa2022facial, pahl2022female, park2022facial, sham2023ethical, hu2025rethinking} and those that further propose methods to mitigate bias \cite{chen2021understanding, cheong2022counterfactual, churamani2022domain, suresh2022using, xu2020investigating, kolahdouzi2023toward}.

Chen and Joo \cite{chen2021understanding}, as well as Suresh and Ong \cite{suresh2022using}, have proposed using facial action units (AUs) to address biases in FER. However, existing evidence suggests that AU recognition models may themselves be biased \cite{pahl2022female, churamani2022domain, fan2021demographic}. Moreover, the lack of demographic labels in public FER datasets has constrained prior studies, limiting the scope of their analysis. For instance, some studies have exclusively focused on a single sensitive attribute, such as age \cite{park2022facial}, gender \cite{domnich2021responsible, manresa2022facial, dominguez2022gender, dominguez2024less}, or race \cite{sham2023ethical}. Others have attempted to analyze multiple sensitive attributes collectively but often restricted their analysis to specific facial expressions, such as happiness \cite{fan2021demographic} or anger \cite{chen2021understanding}. Even research exploring seven basic facial expressions and three sensitive attributes (i.e., age, gender, and race) \cite{xu2020investigating, kolahdouzi2023toward, churamani2022domain, cheong2022counterfactual, cheong2023causal} has been limited to the RAF-DB dataset \cite{li2017reliable}, whose limitations are discussed in Fig.~\ref{fig:1}. In contrast, we provide a more comprehensive bias analysis that includes all three sensitive attributes and seven facial expressions across multiple deep learning models using a large-scale FER dataset.

Annotating perceived demographic labels for large and diverse FER datasets requires significant human effort. Therefore, previous studies have resorted to training facial attribute classifiers to generate pseudo-demographic labels for bias evaluation \cite{chen2021understanding, dominguez2022gender, dominguez2024metrics, dominguez2024less}. However, as shown in our experiments, this approach does not evaluate the biases effectively, particularly for age and race, whereas our proposed feature-level bias evaluation framework addresses this limitation and produces more reliable results. Additionally, prior studies \cite{xu2020investigating, cheong2022counterfactual, churamani2022domain, suresh2022using, hu2025rethinking} have not applied statistical tests for bias evaluation, raising concerns that some reported biases may result from randomness rather than actual disparities. We address this gap by proposing a statistical module to ensure the statistical significance of the reported biases in FER.

\subsection{Evaluating Bias by Association Test}
The concept of association tests, originally derived from the Implicit Association Test (IAT) \cite{greenwald1998measuring,greenwald2003understanding} in psychology, measures biases by having participants rapidly pair concepts with attributes, where shorter response times suggest stronger associations. This approach is commonly used to detect implicit biases in human perception, including those related to how people perceive facial expressions \cite{hammer2015fearful,richetin2004facial,wang2014implicit,steele2018not,lesick2023not}. Building upon this foundation, Caliskan et al. \cite{caliskan2017semantics} adapted the IAT for machine learning, specifically to assess biases within word embeddings, leading to the development of the Word Embedding Association Test (WEAT). 

This approach was extended to the visual domain through the Image Embedding Association Test (iEAT), which measures association biases in computer vision models \cite{steed2021image}, and has also been applied to multimodal models that combine language and vision to examine bias in these systems \cite{wolfe2022american, wolfe2023contrastive}. For example, Sirotkin et al. \cite{sirotkin2022study} conducted a comprehensive study in self-supervised learning visual models, while Brinkmann et al. \cite{brinkmann2023multidimensional} engaged in a multidimensional analysis of social biases in Vision Transformers. However, iEAT is constrained by a fixed set of binary target-attribute pairs and is designed for self-supervised models. In this paper, we reformulate iEAT, tailoring it for multi-class image classification tasks like FER with sensitive attributes involving multiple groups.

\section{Preliminaries}
\subsection{Previous Bias Evaluation Pipeline in FER}
\label{III(A)}

\begin{figure*}[tb]

  \centering
  \captionsetup{justification=centering}
  \includegraphics[height=8cm, width=18cm]{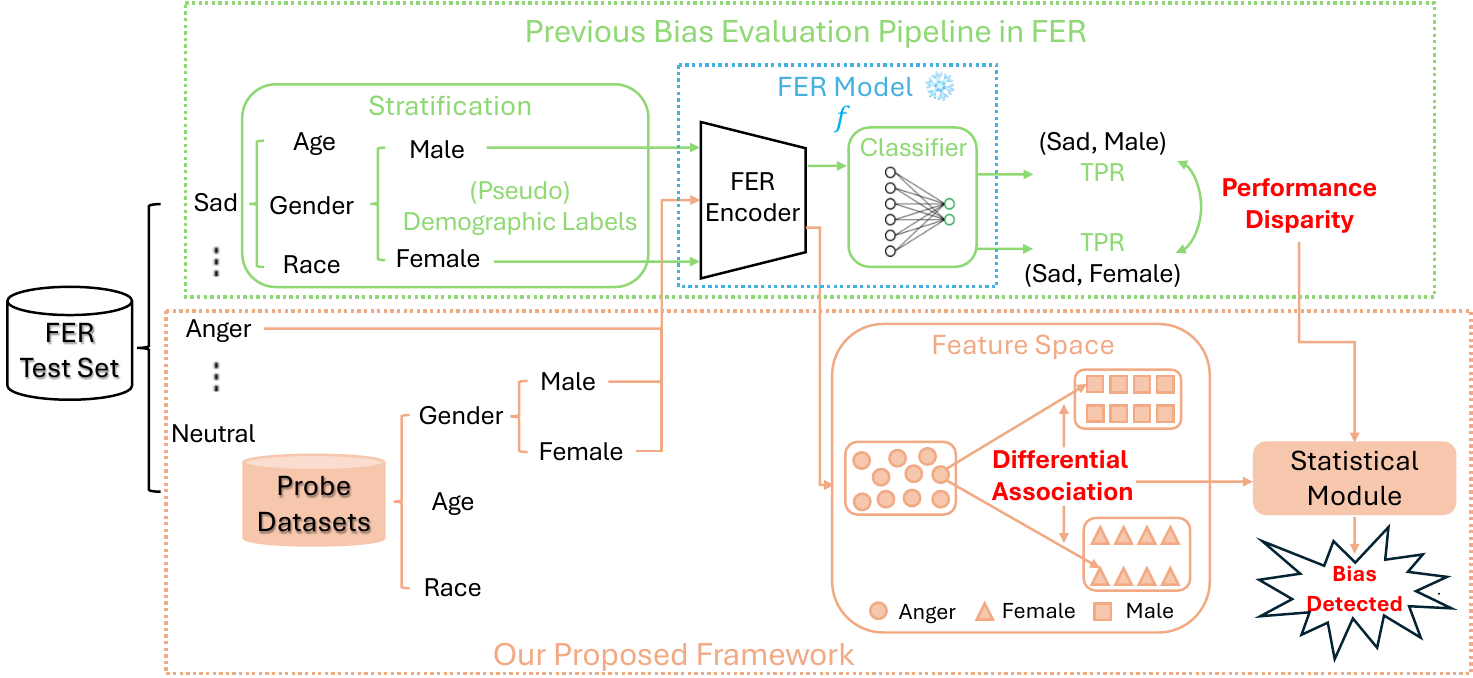}
\caption{Illustration of the previous bias evaluation pipeline in FER (green) and our proposed framework (orange), which evaluates biases in the feature space. A statistical module is also introduced, applicable to both the previous pipeline and our framework, to ensure the statistical significance of observed performance disparities and differential associations. The FER model remains frozen throughout the evaluation.}
  \label{fig:2}
\end{figure*}

The previous bias evaluation pipeline in FER \cite{kolahdouzi2023toward, xu2020investigating, cheong2022counterfactual, churamani2022domain, suresh2022using} for a FER model \(f\) is illustrated in Fig.~\ref{fig:2}. The evaluation is performed on a test set $D^{\mathrm{ts}} = \left\{(x_i,\,y_i,\,s_i)\right\}^n_{i=1}$, where \(x_i\) is a face image, \(y_i\) is its facial expression label, and \(s_i\) is its label or pseudo label for a sensitive attribute \(S\). To examine whether \(f\) behaves differently across each attribute group in $S$ for each facial expression, the test set can be stratified by each facial expression $e \in E$ and $S$:

\begin{equation}
    D^{ts}_{e,S} = \left\{D^{\mathrm{ts}}_{e,\,s_1}, D^{\mathrm{ts}}_{e,\,s_2}, \dots, D^{\mathrm{ts}}_{e,\,s_n}\right\},
\end{equation}

where $S \in \{s_1, s_2, \dots, s_n\}$. We denote $D^{\mathrm{ts}}_{e, s_j}$ as the subset of test set images labeled with facial expression $e$ and an attribute group \(s_j \in S\). For instance, \(D^{\mathrm{ts}}_{\text{anger},\,\text{gender}}\) represents the stratification of test samples labeled anger by gender, with \(D^{\mathrm{ts}}_{\text{anger},\,\text{male}}\) containing samples labeled anger and male while \(D^{\mathrm{ts}}_{\text{anger},\,\text{female}}\) containing samples labeled anger and female. 

We then evaluate $f$ using a performance metric \(M(\cdot)\), such as the true positive rate (TPR), within each \(D^{\mathrm{ts}}_{e, s_j}\). Bias is identified if the disparities in $M(D^{\mathrm{ts}}_{e, s_j})$ across different $s_j$ exceed a predefined threshold. This pipeline has two main limitations.  First, the test set must be equipped with demographic or pseudo-demographic labels to enable stratification. Second, the chosen threshold does not guarantee the statistical significance of the observed performance disparities. As a result, the observed disparities may simply be due to randomness.

\subsection{iEAT}

iEAT \cite{steed2021image} is designed to measure differential associations between pairs of target concepts and attributes using image embeddings. It consists of 15 association tests aimed at evaluating human-like social biases. For instance, one target pair, X and Y, could be flowers vs. insects, while an attribute pair, A and B, could be pleasant vs. unpleasant. The differential association between X, Y, A, and B is defined as:

\begin{equation}
    s(X, Y, A, B) = \sum_{x \in X} s(x, A, B) - \sum_{y \in Y} s(y, A, B),
\end{equation}

\noindent where $s(w, A, B)$ ($w \in \{x, y\}$) represents the differential association of a single image embedding of a target concept  with attributes and is calculated as:

\begin{equation}
   s(w, A, B) = \frac{1}{|A|} \sum_{a \in A} \cos(w, a) - \frac{1}{|B|} \sum_{b \in B} \cos(w, b) 
\end{equation}
\noindent where $a$ and $b$ represent the image embedding for attributes and cos($\cdot$) is the cosine similarity. Statistical tests are then performed to evaluate the significance of the observed associations. However, iEAT is designed for self-supervised models and is limited to a fixed set of 15 binary target–attribute pairs.

\section{Methodology}

\subsection{Evaluating Biases in the Feature Space of FER Models}

Our objective is to evaluate demographic biases in FER models under the setting where demographic labels are unavailable in the test set. Unlike prior FER studies that use pseudo-demographic labels \cite{chen2021understanding, dominguez2022gender, dominguez2024metrics, dominguez2024less}, we propose a method, as shown in Fig.\ref{fig:2}, that uses a "probe" dataset to evaluate biases in the feature space of FER models. A "probe" dataset is a collection of diverse images that represent the visual concepts of interest. Since we aim to evaluate biases in FER models toward specific facial appearances associated with visually perceived demographic groups (e.g., age, gender, and race), we use existing public facial attribute datasets as probe datasets in this paper, as these datasets are specifically designed to cover a broad range of demographic-related facial features. Alternatively, one could also construct this probe dataset from web search engines. 

This approach is further supported by observations from our pilot study, which show that FER models can effectively encode unseen images from different attribute groups across gender, age, and race in the feature space (see Section~\ref{V(C)} for details). Once we have a probe dataset and a test set with facial expression labels, we can evaluate bias in FER models by measuring differential associations in the feature space. Given a facial expression $e$, we feed all images showing facial expression $e$ in the test set into the FER model and extract their embeddings, denoted as $\mathcal{Z}_e$. Similarly, for an attribute group $s_j$ of a sensitive attribute $S$, we feed images labeled as $s_j$ in the probe dataset through the FER model, obtaining a set of embeddings for $s_j$, denoted as $\mathcal{Z}_{s_j}$. By reformulating the Eq. (\textcolor{red}{2}) \& (\textcolor{red}{3}), we compute the association between facial expression $e$ and $s_j$ as:

\begin{equation}
    A(e, s_j) =  \frac{1}{2|\mathcal{Z}_{s_j}| \cdot |\mathcal{Z}_{e}|}  \sum_{z_e \in z_{e}} \sum_{z_{s_j} \in \mathcal{Z}_{s_j}} (\cos(z_e, z_{s_j}) + 1)\end{equation}

\noindent where $z_{e}$ and $z_{s_j}$ represent each embedding in their respective sets and cos($\cdot$) is the cosine similarity. For each facial expression $e$, the differential association (DiA) between two attribute groups $s_j$ and $s_{j^{\prime}}$ is defined as:

\begin{equation}
\text{DiA}^e_{(j, j^{\prime})} \;=\; A(e, s_{j}) \;-\; A(e, s_{j^{\prime}}).
\end{equation}

The intuition is that if the embeddings of the facial expression $e$ are more strongly associated with $s_j$ in the feature space of the FER model being evaluated, then the model is more likely to prefer $s_j$ when predicting $e$. Since any two random sets of embeddings will exhibit some level of differential association, a statistical module is needed to ensure the statistical significance of the observed differential association.

\subsection{Evaluating Biases in FER with a Statistical Module}

Both observed disparities in performance and differential associations require a statistical module to determine whether they are greater than what would occur by chance. Similar to differential association, for each facial expression $e$, we define the disparity in performance (DiP) under a metric $M(\cdot)$ between two attribute groups $s_j$ and $s_{j^{\prime}}$ as:

\begin{equation}
\text{DiP}^{e}_{(j, j^{\prime})} \;=\; M\bigl(D^{\mathrm{ts}}_{e, s_{j}}\bigr) \;-\; M\bigl(D^{\mathrm{ts}}_{e, s_{j^{\prime}}}\bigr).
\end{equation}

For each facial expression $e$ and a sensitive attribute \(S = \{s_1, s_2, \dots, s_n\}\) in the test set, the objective is to identify the attribute groups of $S$ on which the model underperforms and to quantify the extent of this underperformance in a statistically significant manner.  To achieve this, for a given facial expression $e$, we first select the attribute group $s^e_{max}$ with the highest performance:

\begin{equation}
s^{e}_{\max} \;=\; \operatorname*{arg\,max}_{s_j \in \{s_1, s_2, \ldots, s_n\}} M\bigl(D^{\mathrm{ts}}_{e, s_j}\bigr).
\end{equation}

Similarly, in terms of differential association, the attribute group $s^e_{max}$ is the one most strongly associated with facial expression $e$:

\begin{equation}
s^e_{\max} \;=\; \operatorname*{arg\,max}_{s_j \in \{s_1, s_2, \ldots, s_n\}} A\bigl({e, s_j}\bigr).
\end{equation}

For each facial expression $e$, this results in $n-1$ remaining attribute groups, which either underperform or show a weaker association compared to $s^e_{max}$. For each group \( s_k \) among these $n-1$ remaining groups, we set $s^e_{max}$ as the reference group and compute disparities relative to it. The disparity in performance for each $s_k$ is denoted as $\text{DiP}^e_{(max, k)}$ while the differential association is denoted as $\text{DiA}^e_{(max, k)}$, where $ k \in [1, n-1]$. This strategy of comparing each group pairwise to a reference group aligns with the notion of fairness proposed by Feldman et al. \cite{feldman2015certifying}, but with a differential formulation. While it does not capture zero-sum effects between groups, it reflects the pair-wise concept of current legal perspectives on discrimination and is well-suited to sensitive attributes that are either binary or involve multiple groups. Furthermore, this approach ensures that both $\text{DiA}^e_{(max, k)}$ and $\text{DiP}^e_{(max, k)}$ fall within the range [0, 1] with fixed directions. As a result, one-sided $p$-values are sufficient for statistical testing.

To establish a statistical module that applies to both $\text{DiA}^e_{(max, k)}$ and $\text{DiP}^e_{(max, k)}$, we first denote $V^e_k$ as an observed value that represents either $\text{DiA}^e_{(max, k)}$ or $\text{DiP}^e_{(max, k)}$. The goal is to check whether $V^e_k$ could have arisen by chance. To achieve this, we first formulate a null hypothesis $H^e_k$, stating that the observed value $V^e_k$ is purely due to randomness. We then conduct a permutation test for $V^e_k$ under \(H^e_j\). Specifically, we randomly permute the labels between the attribute groups \(s_{max}\) and \(s_k\) while keeping the group sizes unchanged. For each permutation, we obtain a permuted value $V_k^{e,(b)}$ for $b = 1, 2, \dots, B$, where \( B \) is the number of permutations. The one-sided \(p\)-value for each \(V^e_k\), which represents the probability of observing a permuted value $V_k^{e, (b)}$ at least as large as \(V^e_k\) under $H^e_j$, is computed as:
\begin{equation}
    p^e_k = \frac{1}{B}\,\sum_{b=1}^{B}\,\mathbf{1}\!\bigl(V_k^{e, (b)} \ge V^e_k\bigr),
\end{equation}

\noindent where $\mathbf{l}(\cdot)$ is the indicator function. If \( p^e_k < \alpha \) for a chosen significance threshold \( \alpha \) (Type I Error Rate), we reject \( H^e_k \) and conclude that \( V^e_k \) is statistically significant. If \( p^e_k \geq \alpha \), we fail to reject \( H^e_k \), considering \( V^e_k \) as a result of random fluctuation. We then denote $\widetilde{V}^e_k$ as the statistically validated value of \( V^e_k \), where it is retained only if it is statistically significant; otherwise, it is set to zero.

\begin{equation}
    \widetilde{V}^e_k =
    \begin{cases}
        V^e_k, & \text{if } p^e_k < \alpha \\
        0, & \text{otherwise}
    \end{cases}
\end{equation}

\begin{figure*}[h]
    \centering
    \includegraphics[width=15.5cm, height=5cm]{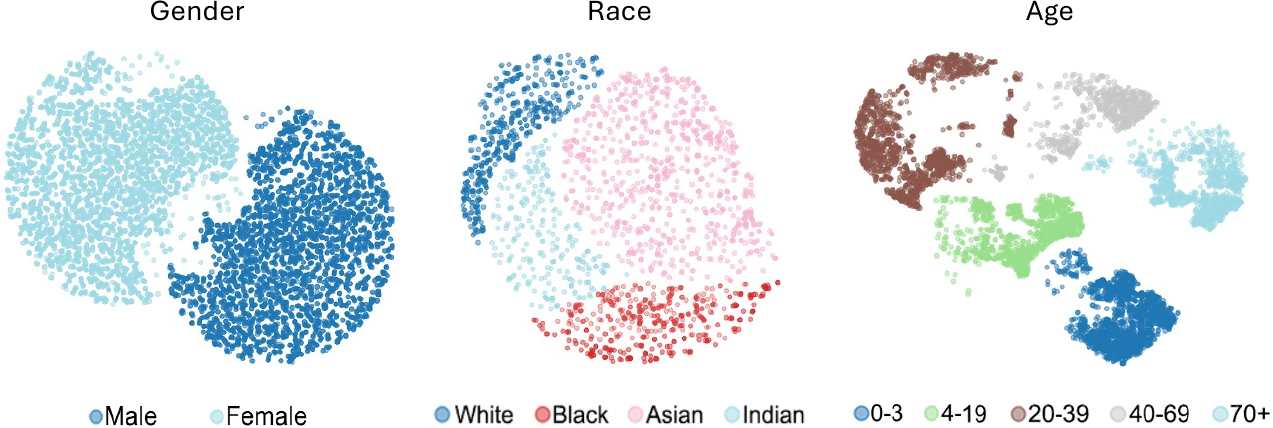}
    \caption{t-SNE visualization from our pilot study: A FER model trained solely on expression labels effectively encodes unseen face images into well-defined clusters based on gender, age, and race, demonstrating that these features are well-preserved in the feature space of FER models.}
    \label{fig:3}
\end{figure*}

\section{Experiments}

\subsection{Datasets}
The datasets used in our study include two FER datasets: RAF-DB and AffectNet, and two facial attribute datasets: FairFace and UTKFace. Among these, RAF-DB and UTKFace are used for the pilot study, while AffectNet, UTKFace, and FairFace are used in the main experiments. All of them are in-the-wild face datasets. We selected in-the-wild datasets because lab-controlled FER datasets often lack participant diversity, making them less suitable for this analysis due to their limited representation of real-world variability and demographic inclusiveness. 

\textbf{RAF-DB} contains in-the-wild face images annotated with facial expressions by 40 expert annotators. Our study focuses on seven basic expressions: neutral, happiness, surprise, sadness, anger, disgust, and fear, resulting in approximately 15,000 face images. \textbf{AffectNet} \cite{mollahosseini2017affectnet} is one of the largest FER datasets and is divided into two subsets: AffectNet-8, which includes eight annotated expressions, and AffectNet-7, which excludes “contempt.” In our experiments, we use AffectNet-7, which contains around 290,000 internet-sourced face images. \textbf{UTKFace} \cite{zhifei2017cvpr} is a facial attribute dataset consisting of 23,704 face images, representing diverse facial appearances across different visually perceived demographic groups. Each image is annotated with visually perceived age, gender, and racial group. \textbf{FairFace} \cite{karkkainen2021fairface} is a facial attribute dataset designed to reduce demographic bias. It contains 108,000 images and provides a balanced distribution of visually perceived gender, age, and racial groups.

\subsection{Data Pre-Processing} 

All face images were aligned and resized to 224×224. For AffectNet, we followed the data-splitting protocol proposed by Hu et al. \cite{hu2025rethinking}, dividing the dataset into training, validation, and independent test sets. Age labels in FairFace and UTKFace were grouped into the same five categories: 0–3, 4–19, 20–39, 40–69, and 70+ years. In UTKFace, images labeled as “others” under racial labels were excluded. For FairFace, race labels were grouped into four categories: White, Black, Asian (East Asian), and Indian. To summarize, the attribute group categorizations used in this paper are as follows: race categories include White, Black, Asian (East Asian), and Indian; age groups are defined as 0 to 3, 4 to 19, 20 to 39, 40 to 69, and 70 years and above; and gender is categorized as Female and Male. Since both facial expression and facial attribute labels are subjective, we identified visually ambiguous samples that were likely misannotated across the datasets we used. These samples were excluded from our analysis. A list of the excluded images is provided in the supplementary material to support reproducibility. For more details on the data cleaning process, please refer to the \textbf{Supplementary Material}.

\subsection{Pilot Study: FER Models Are Good Attribute Encoders}
\label{V(C)}

FER models are naturally effective facial attribute encoders. This is because, although they are not explicitly trained to recognize sensitive attributes, these features are inherently present in face images and are therefore indirectly encoded during training. We conduct a small pilot study as a proof of concept. First, we train a FER model based on ResNet34 using the RAF-DB dataset with only facial expression labels. We then keep the trained model frozen and use it to encode images from the UTKFace dataset. As shown in Fig.~\ref{fig:3}, the resulting image embeddings form distinct clusters corresponding to gender, race, and age. This supports that differential associations in our framework can be effectively computed in the feature space of the FER models being evaluated.

\subsection{Settings for the Main Experiments}

In our main experiments, we use AffectNet to ensure sufficient sample size and statistically significant results, as it is nearly 20 times larger than RAF-DB. Specifically, we train FER models on the AffectNet training and validation sets and evaluate bias on the test set. Furthermore, we implement two versions of our proposed framework using UTKFace and FairFace as probe datasets, respectively, to examine the sensitivity of our approach to probe datasets of different scales (FairFace is nearly four times larger than UTKFace). For a fair comparison with existing fair FER studies \cite{chen2021understanding, dominguez2022gender, dominguez2024metrics, dominguez2024less} that rely on pseudo-demographic labels, we implement them by training facial attribute classifiers using the same datasets. Please refer to the \textbf{Supplementary Material} for implementation details. These classifiers are then used to generate pseudo-demographic labels for the AffectNet test set. The generated pseudo-labels are used to conduct bias evaluation. This setup results in four bias evaluation methods for comparison in our experiments:

1. Bias evaluation using pseudo-demographic labels generated by an attribute classifier trained on UTKFace, denoted as UTKFace (pseudo).

2. Bias evaluation using pseudo-demographic labels generated by an attribute classifier trained on FairFace, denoted as FairFace (pseudo).

3. Bias evaluation using UTKFace as a probe dataset to compute differential associations, denoted as UTKFace (ours).

4. Bias evaluation using FairFace as a probe dataset to compute differential associations, denoted as FairFace (ours).

\subsubsection{Implementation Details}
The FER models we train and evaluate include ResNet18, ResNet34, ResNet101, and ResNet152, which represent convolutional neural networks (CNNs) \cite{he2016deep}, and Vision Transformer Base, Swin Transformer Tiny, Swin Transformer Small, and Swin Transformer Base, which represent Transformer-based architectures \cite{liu2021swin, dosovitskiy2020image}. We choose these architectures because they are among the most commonly used backbones for feature extraction in FER. All models are optimized using stochastic gradient descent (SGD) with the same hyperparameters. Feature embeddings are extracted from the penultimate layer, with a dimension of 512. In this paper, we present detailed results using the Swin Transformer (Base) \cite{liu2021swin} as a representative architecture, chosen for its strong performance in FER and related areas \cite{hu2025rethinking, luo2022learning, lian2023supervised}.  Please refer to the \textbf{Supplementary Material} for the training details and the results of other models. 

We use the true positive rate (TPR) as the performance metric $M(\cdot)$, which aligns with the concept of Equal Opportunity, a fairness metric widely adopted in prior studies on fairness in FER \cite{kolahdouzi2023toward, xu2020investigating, cheong2022counterfactual, churamani2022domain, suresh2022using}. In this context, $\text{DiP}_{(max, k)}^e$ represents the difference in Equal Opportunity, which we refer to as $\text{DEO}_{(max, k)}^e$ for the remainder of the paper. Following prior research in iEAT \cite{sirotkin2022study, steed2021image, brinkmann2023multidimensional, cohen2013statistical}, we generate 10,000 permutations for each test ($B = 10000$) and set the significance threshold in the statistical module to $\alpha = 0.05$. All implementations are carried out using PyTorch and two NVIDIA A100 GPUs.

\subsubsection{Evaluating the Bias Evaluation Methods}

To evaluate the effectiveness of the four aforementioned bias evaluation methods, we use the bias evaluation results based on human-annotated demographic labels provided by Hu et al. \cite{hu2025rethinking} on the AffectNet test set to serve as the ground truth bias evaluation results. If a method $\mathcal{M}$, which operates under the setting where demographic labels are unavailable in the test set, produces results closely aligned with those based on human annotations, it indicates that $\mathcal{M}$ is more effective. We pass the bias evaluation results from the four bias evaluation methods, along with the ground truth, through our proposed statistical module to ensure that all the reported biases are not the result of random fluctuations. 

To quantify the closeness, for each facial expression $e$ and a sensitive attribute $S$ with $n$ groups, let $s^{e, \mathcal{M}}_{max}$ be the reference group identified by  $\mathcal{M}$ as having the highest performance or strongest association with $e$, and let $s^{e, GT}_{max}$ be the corresponding group in the ground truth. If $s^{e, \mathcal{M}}_{max} = s^{e, GT}_{max}$, we compute the average $L_1$ distance between the statistically validated values $\widetilde{V}^{e, \mathcal{M}}_k$ produced by $\mathcal{M}$ and the corresponding ground truth values $\widetilde{V}^{e, GT}_k$ for the remaining $n-1$ groups, as follows: 

\begin{equation}
    L_1^{e, \mathcal{M}} = \frac{1}{n-1}\sum\limits_{k=1}^{n-1}{|\widetilde{V}^{e, \mathcal{M}}_k - \widetilde{V}^{e, GT}_k|},
\end{equation}

\noindent where $\widetilde{V}^{e, \mathcal{M}}_k$ is derived from either $\text{DEO}^{e, \mathcal{M}}_{(max, k)}$ or $\text{DiA}^{e, \mathcal{M}}_{(max, k)}$, and $\widetilde{V}^{e, GT}_k$ is derived from $\text{DEO}^{e, GT}_{(max, k)}$. Since both $\widetilde{V}^{e, \mathcal{M}}_k$ and $\widetilde{V}^{e, GT}_k$ lie within the range $[0, 1]$, the resulting distance $L_1^{e, \mathcal{M}}$ also falls within the interval $[0, 1]$. If $s^{e, \mathcal{M}}_{max} \neq s^{e, GT}_{max}$,  then all values $\widetilde{V}^{e, \mathcal{M}}_k$ from $\mathcal{M}$ are computed with respect to an incorrect reference group compared to the ground truth, which means $\mathcal{M}$ fails entirely for this expression. In such cases, we do not compute the $L_1^{e, \mathcal{M}}$ and mark the result as $\text{NaN}$.

\begin{table*}[ht]
\centering
\caption{Ground truth demographic bias across seven facial expressions for gender. If there is no statistically significant bias between the reference group and the remaining group, the relationship is denoted as $\{\text{reference group}\}$/$\{\text{remaining group(s)}\}$. Otherwise, it is denoted as $\{\text{reference group}\}$ $\rightarrow$ $\{\text{remaining group(s)}\}$.  $\widetilde{V}^{e, GT}_k$ values are reported as percentages. (Female: F, Male: M)}
\setlength{\tabcolsep}{4pt}
\renewcommand{\arraystretch}{1.4}
{\small%
\begin{tabular}{c|c|c|c|c|c|c|c}
\hline
\multicolumn{1}{c|}{Expressions} & Surprise & Fear & Disgust & Happiness & Sadness & Anger & Neutral \\ \hline
Bias Direction & F/M & F/M & F $\rightarrow$ M & F $\rightarrow$ M & F $\rightarrow$ M & M $\rightarrow$ F & F $\rightarrow$ M \\ \hline
$\widetilde{V}^{e, GT}_k$ (\%) & 0 & 0 & 9.01 & 4.94 & 3.22 & 9.30 & 2.41 \\ \hline
\end{tabular}%
}
\label{table:I}
\end{table*}

\begin{table*}[ht]
\centering
\caption{Comparison of bias evaluation methods across seven facial expressions for gender. $L_1^{e, \mathcal{M}}$ values are reported as percentages. A lower $L_1^{e, \mathcal{M}}$ indicates better alignment with the ground truth. Red highlights mark the method that achieves the lowest value for each expression. If a method identifies the incorrect reference group, the result is marked as NaN.}
\setlength{\tabcolsep}{4pt}
\renewcommand{\arraystretch}{1.3}
{\small%
\begin{tabular}{c|c|c|c|c|c|c|c}
\hline
\multirow{2}{*}{\diagbox{Method}{Expression}} & Surprise & Fear & Disgust & Happiness & Sadness & Anger & Neutral \\ \cline{2-8}
 & \multicolumn{7}{c}{$L_1^{e, \mathcal{M}}$ (\%) \textcolor{red}{$\downarrow$}} \\ \hline
FairFace (ours) & 0 & 0 & \cellcolor{red!40}0.23 & 0.22 & 0.24 & \cellcolor{red!40}0.31 & 0.14 \\ \hline
UTKFace (ours) & 0 & 0 & 0.42 & \cellcolor{red!40}0.13 & 0.32 & 0.33 & \cellcolor{red!40}0.12 \\ \hline
FairFace (pseudo) & 0 & 0 & 1.24 & 1.53 & \cellcolor{red!40}0.41 & 0.82 & 1.92 \\ \hline
UTKFace (pseudo) & 0 & 0 & 1.43 & 1.91 & 1.24 & 1.52 & 2.21 \\ \hline
\end{tabular}%
}
\label{table:II}
\end{table*}

\begin{table*}[!ht]
\centering
\caption{Ground truth demographic bias across seven facial expressions for the sensitive attribute race. If there is no statistically significant bias between the reference group and the remaining group(s), the relationship is denoted as $\{\text{reference group}\}$/$\{\text{remaining group(s)}\}$. Otherwise, it is denoted as $\{\text{reference group}\}$ $\rightarrow$ $\{\text{remaining group(s)}\}$. $\widetilde{V}^{e, GT}_k$ values are reported as percentages. Blue values indicate $\widetilde{V}^{e, GT}_k$ greater than 10\%. (White: W, Black: B, Asian: A, Indian: I)}
\setlength{\tabcolsep}{2pt}
\renewcommand{\arraystretch}{1.4}
{\small
\begin{tabular}{c|c|c|c|c|c|c|c}
\hline
\multicolumn{1}{c|}{Expressions} & Surprise & Fear & Disgust & Happiness & Sadness & Anger & Neutral \\ \hline
Bias Direction & W $\rightarrow$ (B, A, I) & I $\rightarrow$ (W, B, A) & W/B/A $\rightarrow$ I & W/B/A/I & A $\rightarrow$ (W, B, I) & W/I/A $\rightarrow$ B & B/A $\rightarrow$ (W, I) \\ \hline
$\widetilde{V}^{e, GT}_k$ (\%) & (8.23, \textcolor{blue}{10.92}, 9.63) & (7.24, 5.62, \textcolor{blue}{16.63}) & 5.72 & 0.00 & (8.23, \textcolor{blue}{10.93}, \textcolor{blue}{11.82}) & 7.81 & (4.23, 5.82) \\ \hline
\end{tabular}%
}
\label{table:III}
\end{table*}

\subsection{Experimental Results}
In this section, we present the bias evaluation results of the four bias evaluation methods and the ground truth across gender, race, and age. We report $\widetilde{V}^{e, GT}_k$  and $L_1^{e, \mathcal{M}}$ as percentages since they both fall within the range [0, 1]. We note that a value of $\widetilde{V}^{e, GT}_k = 0$ does not indicate that the observed value is zero. Instead, it means that the observed value is not statistically significant, as we set it to zero when $p \ge 0.05$ to simplify the presentation of our experimental results. Similarly, $L_1^{e, \mathcal{M}} = 0$ does not imply that the results of $\mathcal{M}$ perfectly match the ground truth. It simply indicates that $\mathcal{M}$ also found the result to be not statistically significant. For detailed values of $\widetilde{V}^{e, \mathcal{M}}_k$ and the corresponding $p$-values across facial expressions and sensitive attributes for each method, please refer to the \textbf{Supplementary Material}.

\subsubsection{Gender} Our analysis of gender is shown in Table~\ref{table:I} and Table~\ref{table:II}. As shown in Table~\ref{table:I}, the ground truth results show no statistically significant gender bias for the expressions of surprise and fear. However, the FER model performs better on female-looking faces for disgust, happiness, sadness, and neutral. In contrast, for anger, it shows a preference for male-looking faces. As shown in Table~\ref{table:II}, all four evaluation methods correctly capture these trends as there are no NaN values. When comparing the exact values, for disgust, happiness, anger, and neutral, our proposed methods using UTKFace and FairFace (ours) are closer to the ground truth than UTKFace and FairFace (pseudo), as measured by the $L_1$ distance. 

Our findings on gender bias in happiness resonate with previous research showing that females are often perceived as generally happier than males \cite{chen2021understanding, manresa2022facial, dominguez2022gender}. This is further supported by studies indicating that females typically exhibit higher AU12 (smile muscle) intensity than males \cite{fan2021demographic}, and psychological investigations using the IAT \cite{richetin2004facial}, which suggest that cosmetics may also play a role. Similarly, for anger, earlier studies also report that FER models tend to favor male-looking faces \cite{manresa2022facial, chen2021understanding}.

\begin{table*}[ht]
\centering
\caption{Comparison of bias evaluation methods across seven facial expressions for race. $L_1^{e, \mathcal{M}}$ values are reported as percentages. A lower $L_1^{e, \mathcal{M}}$ indicates better alignment with the ground truth. Red highlights indicate the method that achieves the lowest value for each expression. If a method $\mathcal{M}$ identifies the incorrect reference group, the result is marked as NaN.}
\setlength{\tabcolsep}{4pt}
\renewcommand{\arraystretch}{1.4}
{\small
\begin{tabular}{c|c|c|c|c|c|c|c}
\hline
\multirow{2}{*}{\diagbox{Method}{Expression}} & Surprise & Fear & Disgust & Happiness & Sadness & Anger & Neutral \\ \cline{2-8}
 & \multicolumn{7}{c}{$L_1^{e, \mathcal{M}}$ (\%) \textcolor{red}{$\downarrow$}} \\ \hline
FairFace (ours) & 0.89 & \cellcolor{red!40}0.12 & \cellcolor{red!40}0.26 & 0.00 & 0.84 & \cellcolor{red!40}0.08 & \cellcolor{red!40}0.15 \\ \hline
UTKFace (ours) & \cellcolor{red!40}0.71 & 0.89 & 0.40 & 0.00 & \cellcolor{red!40}0.56 & 0.44 & 0.47 \\ \hline
FairFace (pseudo) & 4.95 & NaN & 3.17 & 0.00 & NaN & 5.99 & 4.11 \\ \hline
UTKFace (pseudo) & NaN & NaN & 3.23 & 0.00 & NaN & 5.71 & 5.37 \\ \hline
\end{tabular}%
}
\label{table:IV}
\end{table*}

\begin{table*}[ht]
\centering
\caption{Ground truth demographic bias across seven facial expressions for age. If there is no statistically significant bias between the reference and remaining group(s), the relationship is denoted as $\{\text{reference group}\}$/$\{\text{remaining group(s)}\}$. Otherwise, it is denoted as $\{\text{reference group}\}$ $\downarrow$ $\{\text{remaining group(s)}\}$. $\widetilde{V}^{e, GT}_k$ values are reported as percentages. Blue values indicate $\widetilde{V}^{e, GT}_k$ greater than 20\%. ($\text{A}_1$: 0–3, $\text{A}_2$: 4–19, $\text{A}_3$: 20–39, $\text{A}_4$: 40–69, $\text{A}_5$: 70+ years)}
\setlength{\tabcolsep}{0.5pt}
\renewcommand{\arraystretch}{1.4}
{\fontsize{8pt}{9.8pt}\selectfont
\begin{tabular}{c|c|c|c|c|c|c}
\hline
\ Surprise & Fear & Disgust & Happiness & Sadness & Anger & Neutral \\ \hline
  $\text{A}_1$/$\text{A}_5$ & $\text{A}_1$ & $\text{A}_2$/$\text{A}_3$ & $\text{A}_1$/$\text{A}_2$/$\text{A}_3$/$\text{A}_5$ & $\text{A}_1$ & $\text{A}_1$/$\text{A}_2$ & $\text{A}_2$/$\text{A}_3$ \\ [-5pt]
 $\downarrow$ & $\downarrow$ & $\downarrow$ & $\downarrow$ & $\downarrow$ & $\downarrow$ & $\downarrow$ \\ [-5pt]
  ($\text{A}_2$, $\text{A}_3$, $\text{A}_4$) & ($\text{A}_2$,$\text{A}_3$, $\text{A}_4$, $\text{A}_5$) & ($\text{A}_1$, $\text{A}_4$, $\text{A}_5$) & $\text{A}_4$ & ($\text{A}_2$, $\text{A}_3$, $\text{A}_4$, $\text{A}_5$) & ($\text{A}_3$, $\text{A}_4$, $\text{A}_5$) & ($\text{A}_1$, $\text{A}_4$, $\text{A}_5$) \\ \hline
  \multicolumn{7}{c}{$\widetilde{V}^{e, GT}_k$ (\%)} \\ \hline 
 (13.24, 15.31, 14.23) & (13.34, 14.13, 14.42, \textcolor{blue}{25.31}) & (14.32, 17.14, \textcolor{blue}{34.23}) & 3.14 & (17.23, \textcolor{blue}{33.14}, \textcolor{blue}{31.42}, \textcolor{blue}{28.13}) & (\textcolor{blue}{21.31}, 14.23, \textcolor{blue}{30.14}) & (13.42, 14.31, \textcolor{blue}{25.24}) \\ \hline
\end{tabular}
}
\label{table:V}
\end{table*}

\begin{table*}[ht]
\centering
\caption{Comparison of bias evaluation methods across seven facial expressions for the sensitive attribute age. $L_1^{e, \mathcal{M}}$ values are reported as percentages. A lower $L_1^{e, \mathcal{M}}$ indicates better alignment with the ground truth. Red highlights mark the method that achieves the lowest value for each expression. If a method identifies the incorrect reference group, the result is marked as NaN.}
\setlength{\tabcolsep}{4pt}
\renewcommand{\arraystretch}{1.4}
{\small
\begin{tabular}{c|c|c|c|c|c|c|c}
\hline
\multirow{2}{*}{\diagbox{Method}{Expression}} & Surprise & Fear & Disgust & Happiness & Sadness & Anger & Neutral \\ \cline{2-8}
 & \multicolumn{7}{c}{$L_1^{e, \mathcal{M}}$ (\%) \textcolor{red}{$\downarrow$}} \\ \hline
FairFace (ours) & 0.44 & 0.42 & \cellcolor{red!40}0.41 & 0.49 & \cellcolor{red!40}0.45 & \cellcolor{red!40}0.38 & \cellcolor{red!40}0.37 \\ \hline
UTKFace (ours) & \cellcolor{red!40}0.31 & \cellcolor{red!40}0.35 & 0.58 & \cellcolor{red!40}0.30 & 0.52 & 0.51 & 0.46 \\ \hline
FairFace (pseudo) & NaN & 5.12 & 5.83 & 5.50 & NaN & NaN & 5.94 \\ \hline
UTKFace (pseudo) & NaN & NaN & NaN & 8.00 & NaN & NaN & 8.83 \\ \hline
\end{tabular}
}
\label{table:VI}
\vspace{-3pt}
\end{table*}

\subsubsection{Race} Our analysis of racial bias is summarized in Table~\ref{table:III} and Table~\ref{table:IV}. As shown in Table~\ref{table:III}, the ground truth results indicate no statistically significant bias across racial groups for the expression of happiness. For surprise, the FER model predicts White faces more accurately than Black, Asian, and Indian faces. For fear, the model performs best on Indian faces, while prediction accuracy is lower for other groups. In the case of disgust, there is no statistically significant bias among White, Black, and Asian groups, all of which outperform the Indian group. For sadness, Asian faces are predicted more accurately than all other groups. For anger, there is no statistically significant bias for White, Indian, and Asian groups, but the model underperforms on Black faces. For the neutral expression, no statistically significant bias is found for Black and Asian groups, whereas White and Indian groups are predicted less accurately.

As shown in Table~\ref{table:IV}, UTKFace (pseudo) fails to identify the reference group for surprise, fear, and sadness, resulting in NaN values for these expressions. Similarly, FairFace (pseudo) fails to identify the reference group for fear and sadness, also producing NaN values. In contrast, both UTKFace and FairFace (ours) correctly identify the reference group across all expressions. When comparing exact values, UTKFace (ours) is closest to the ground truth for surprise and sadness, while FairFace (ours) aligns most closely with the ground truth for fear, disgust, sadness, anger, and neutral. Our finding for the anger expression echoes psychological research showing that people are more likely to associate Black faces with hostility \cite{hugenberg2004ambiguity} or anger \cite{lesick2023not, steele2018not}. This pattern is also observed by Rhue et al. \cite{rhue2018racial}, who reported that modern FER APIs more frequently classified Black NBA players as angry compared to their White counterparts.

\subsubsection{Age} Our analysis of age bias is presented in Table~\ref{table:V} and Table~\ref{table:VI}. As shown in Table~\ref{table:V}, for the expression surprise, the ground truth indicates no statistically significant bias for the 0–3 and 70+ age groups, while other groups are predicted less accurately. For fear, the model performs best on the 0–3 age group, with all other groups showing lower accuracy. For disgust, the model predicts the 4–39 age range more accurately than other groups. For happiness, the model underperforms only for the 40–69 age group, with no statistically significant bias for the remaining groups. For sadness, the model favors the 0–3 age group, with all others performing worse by comparison. For anger, the model favors the 0–19 age group, and prediction accuracy decreases for older groups. The model shows the highest accuracy for the neutral expression for the 4–39 age range, while other groups are predicted less accurately.

As shown in Table~\ref{table:VI}, UTKFace (pseudo) fails to identify the reference group for surprise, fear, disgust, sadness, and anger, resulting in NaN values for these expressions. Similarly, FairFace (pseudo) fails to identify the reference group for surprise, sadness, and anger, also producing NaN values. In contrast, both UTKFace and FairFace (ours) correctly identify the reference group across all expressions. When comparing the exact values, UTKFace (ours) is closest to the ground truth for surprise, fear, and happiness, while FairFace (ours) aligns best with the ground truth for disgust, sadness, anger, and neutral. Our finding for the fear expression aligns with psychological research using the Implicit Association Test (IAT) \cite{hammer2015fearful} to measure implicit biases in human perception of facial expressions, which suggests an implicit association between fearful expressions and infant faces.

\subsubsection{Summary}
This section highlights key takeaways and observations in our experiments from Tables~\ref{table:I} to~\ref{table:VI}, which are also consistent across all evaluated network architectures.

\textbf{Biases are more pronounced in age and race than in gender.} Based on the ground truth results shown in Table~\ref{table:I}, Table~\ref{table:III}, and Table~\ref{table:V}, which report fairness across gender, race, and age, we observe that several $\widetilde{V}^{e, \mathcal{M}}_k$ values for age exceed 20\%, and some values for race exceed 10\%. In contrast, all $\widetilde{V}^{e, \mathcal{M}}_k$ values for gender remain below 10\%. This discrepancy may be because race and age categories include multiple subgroups, making it more difficult to create a balanced training dataset, whereas gender typically involves only two groups.

\textbf{Methods based on pseudo-demographic labels perform worse on age and race.} As shown in Table~\ref{table:II}, Table~\ref{table:IV}, and Table~\ref{table:VI}, methods based on pseudo-labels tend to be more error-prone and less reliable for race and age compared to gender, often resulting in a higher number of NaN values and larger $L_1$ distances from the ground truth. We attribute this to the high accuracy of facial attribute classifiers in distinguishing male and female faces, which makes gender-based pseudo-labels more reliable. However, the performance of these classifiers for race and age is substantially lower compared to gender. The reduced accuracy in predicting racial and age groups makes bias evaluation based on unreliable pseudo-labels and may skew the bias evaluation results. In contrast, our proposed method performs consistently well across gender, race, and age. We attribute this to the fact that our method evaluates bias directly in the feature space of the FER model being evaluated. This space inherently reflects how the model encodes and separates input features, allowing us to capture biased behavior more effectively.

\textbf{Methods based on pseudo-demographic labels are more sensitive to the facial attribute datasets.} We also observe that the performance of these methods varies depending on the dataset used to train the facial attribute classifier. For example, FairFace, which contains nearly four times more samples than UTKFace, results in better bias evaluation performance in these methods. It produces fewer NaN values and lower $L_1$ scores because the pseudo-demographic labels are more accurate when trained on a larger dataset. In comparison, our method yields similarly strong bias evaluation results using either UTKFace or FairFace as the probe dataset. This indicates that a smaller dataset like UTKFace is already sufficient for producing reliable bias evaluations for FER models in our framework. Unlike methods based on pseudo labels, our approach depends less on large-scale data to be effective.

\textbf{Similar patterns of biases are observed across studies.} As discussed in previous sections, several of our findings are consistent with results reported in prior studies on demographic bias in FER models, as well as in psychological research. Although our experiments are conducted under different settings, these similarities may indicate that such biases are widespread in FER datasets, frequently emerge during the annotation process, or are further amplified by deep learning models. This underscores the importance of mitigating these biases with greater care.

    \begin{figure}[!t]
      \centering
      \includegraphics[height=6cm,width=8.5cm]{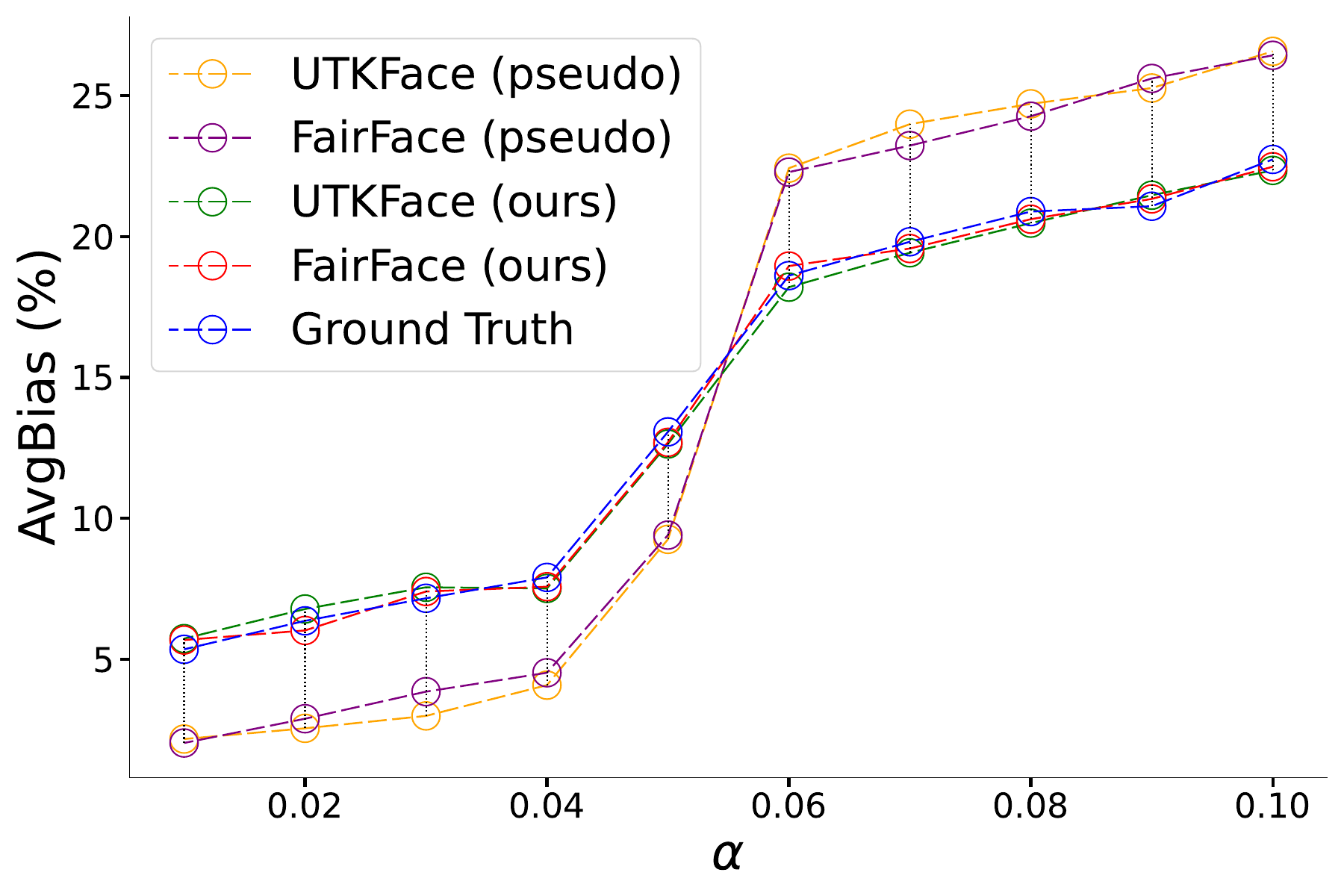}

      \caption{Sensitivity analysis of the threshold $\alpha$ across four bias evaluation methods. The closer the curves align with the ground truth, the better the evaluation method.}
       \label{fig:4}
    
    \end{figure}

\subsection{A Sensitivity Analysis of Threshold $\alpha$}

As shown in Fig.~\ref{fig:4}, we conducted a sensitivity analysis on the threshold $\alpha$ used in our statistical module. For each bias evaluation method $\mathcal{M}$ and the ground truth, we compute the average of $\widetilde{V}^{e, \mathcal{M}}_k$ and $\widetilde{V}^{e, GT}_k$ across the $n-1$ underperforming or less associated groups, the seven facial expressions, and the three sensitive attributes. This yields a single average value, defined as:
\begin{equation}
    \text{AvgBias} = \frac{1}{(n-1)|E||\mathcal{S}|}\sum\limits_{S \in \mathcal{S}}\sum\limits_{e \in E}\sum\limits_{k=1}^{n-1}{\widetilde{V}_k^{e,S}}
\end{equation}

\noindent where each sensitive attribute $S \in \mathcal{S} = \{\text{gender}, \text{race}, \text{age}\}$ and $\widetilde{V}_k^{e,S}$ represents either $\widetilde{V}^{e, \mathcal{M}}_k$ or $\widetilde{V}^{e, GT}_k$ under the sensitive attribute $S$. NaN values are excluded from the calculations. This average reflects the overall level of bias across underperforming attribute groups, sensitive attributes, and facial expressions. We analyze how this average changes as the threshold $\alpha$ varies from 0.01 to 0.1 in increments of 0.01.

Intuitively, as $\alpha$ decreases toward 0.01, the criterion for statistical significance becomes more stringent. Consequently, more observed $\widetilde{V}_k^{e,S}$ values become statistically insignificant and are set to zero, leading to a decrease in $\text{Avg}{\text{Bias}}$, as shown in Fig.~\ref{fig:4}. Conversely, as $\alpha$ increases toward 0.1, the criterion becomes less stringent, allowing more values to be considered statistically significant and increasing $\text{AvgBias}$. Fig.~\ref{fig:4} also shows that our proposed method consistently follows the trend of the ground truth across different values of $\alpha$. In contrast, methods based on pseudo-labels show larger deviations from the ground truth compared to our method.

\subsection{Exploring More Network Architectures}

\begin{figure}[!t]
      \centering
      \includegraphics[height=6cm,width=8.5cm]{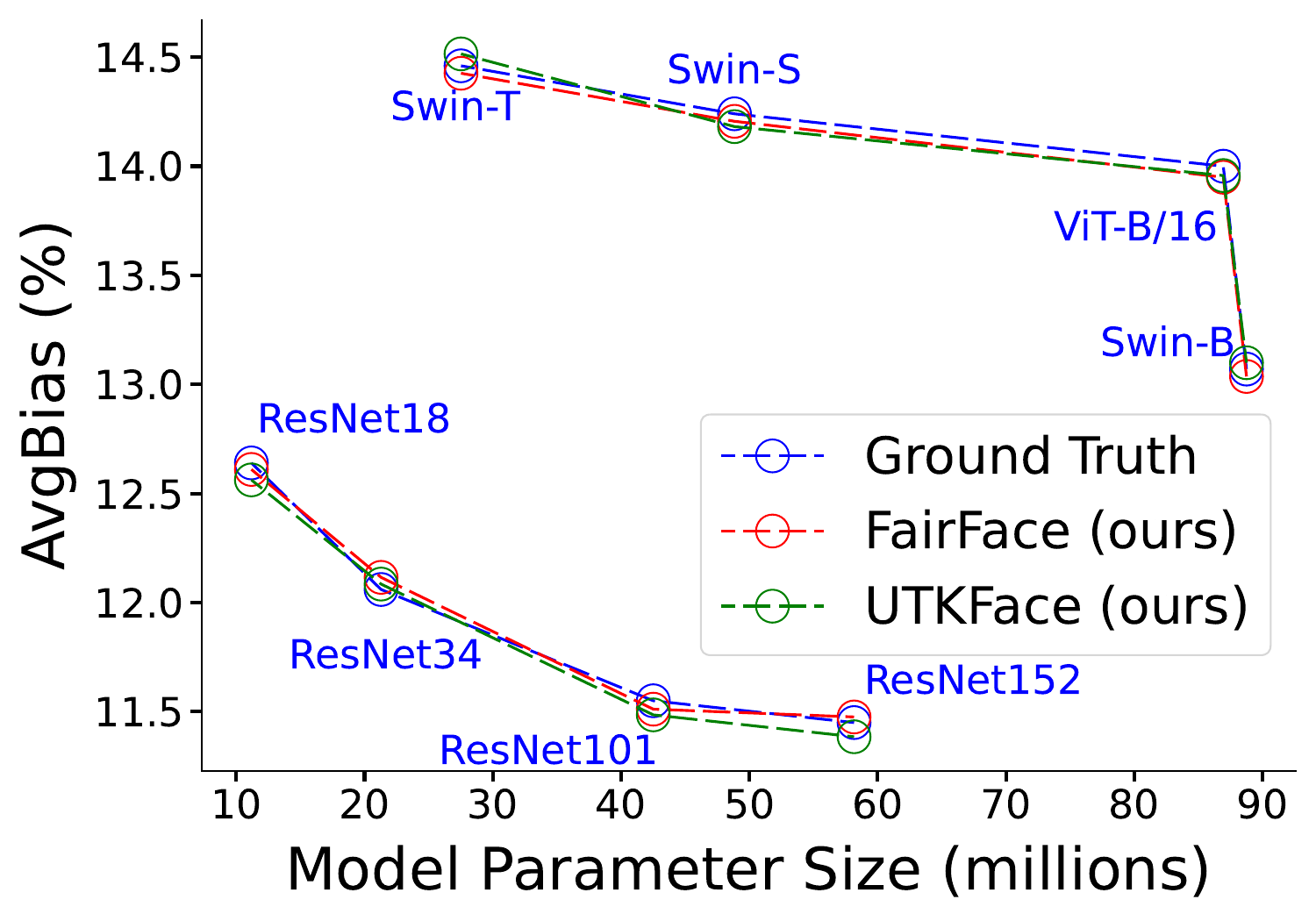}

      \caption{Experimental results of our proposed methods across multiple network architectures.}
            \label{fig:5}
    
    \end{figure}

As shown in Fig.\ref{fig:5}, we evaluate the effectiveness of our bias evaluation methods across eight network architectures based on the $\text{AvgBias}$. All models were trained using the same set of hyperparameters to ensure that any differences in $\text{AvgBias}$ are due solely to architectural differences. The network architectures are ordered by model size. As shown in Fig.\ref{fig:5}, our proposed methods remain highly consistent with the ground truth across both CNN and Transformer families. Our results also indicate that Transformer-based models exhibit higher FER bias compared to their CNN-based counterparts. A similar observation is noted by Mandal et al. \cite{mandal2023biased}, who reported that Vision Transformers tend to amplify gender bias. We believe this may be due to their global attention mechanism and the larger receptive fields introduced by multi-headed self-attention, which can lead the models to capture broader visual attributes such as race, age, and gender, rather than focusing primarily on localized facial action units or expressions. In addition, we observe that bias tends to decrease as model size increases for both the ResNet and Transformer families. A similar trend was reported by Brinkmann et al. \cite{brinkmann2023multidimensional}, who suggested that larger models are better at capturing the core semantic content of images, which may reduce reliance on sensitive attributes.

\section{Discussion}

\subsection{On the Use of Visually Perceived Demographic Labels}
\label{VI(A)}

Image-level demographic annotations are essential for advancing fairness research in computer vision. However, in large-scale datasets such as the "people" subtree in ImageNet \cite{yang2020towards}, these annotations are often based on perceived attributes rather than self-identified demographics. It is important to note that these two types of information are fundamentally different. While self-reported data is highly valuable, collecting it at scale is extremely difficult, as it would require participation from a large number of volunteers. Consequently, much of the fairness research in computer vision \cite{chen2021understanding, cheong2022counterfactual, dominguez2024metrics, xu2020investigating} relies on annotations derived from the perceptions of external annotators. Evaluating fairness based on perceived attributes is an important step for computer vision systems that rely solely on image-based inputs. For FER models, consistent performance across facial appearances associated with perceived gender, race, and age is essential to prevent discrimination against demographic-related facial features in downstream applications, thereby supporting the responsible development of this technology.

\section{Conclusion}

In this paper, we propose a feature-level bias evaluation framework to evaluate demographic biases in FER models in scenarios where demographic labels are not available in the test set. Extensive experiments demonstrate that our method more effectively evaluates demographic biases compared to existing approaches that rely on pseudo-demographic labels. To ensure the statistical significance of the evaluation results, we introduce a plug-and-play statistical module that can be integrated into both existing evaluation pipelines and our proposed method that analyzes the feature space. Furthermore, a comprehensive bias analysis based on the statistical module is then conducted across three sensitive attributes (age, gender, and race), seven facial expressions, and multiple network architectures on a large-scale dataset, revealing the prominent demographic biases in FER and providing insights on selecting a fairer network architecture.

An additional strength of our method is its ability to evaluate biases embedded in the feature space of FER models. Identifying such biases is important, as they can extend beyond FER. For example, biases in the feature space may propagate to downstream tasks during fine-tuning \cite{wang2023overwriting} or through feature fusion in multi-modal emotion recognition systems \cite{zhang2023transformer}. In future work, we plan to investigate how these feature-level biases transfer to other tasks through fine-tuning and feature fusion, and to develop mitigation strategies that prevent this transfer of bias.

\section{Ethical Statement}

Demographic categories are socially constructed, context-dependent, and continue to be the focus of ethical and academic debate regarding their appropriateness and feasibility in computational systems \cite{keyes2018misgendering, hamidi2018gender, larson2017gender}. In this work, demographic categories are used only as descriptors of facial appearance, rather than as indicators of biological sex, self-identified gender, race, or age of the individual. These labels are used solely to examine how FER models respond to different facial characteristics. This work does not support or endorse the use of such annotations to determine or assign the actual gender or race of any person. There is a clear ethical responsibility to treat demographic information with care, and this study is committed to advancing inclusive and respectful practices in the community.

\bibliographystyle{IEEEtran}

\bibliography{mypaper}

 




\vfill

\newpage


\section*{Supplementary material}

# Supplementary Material: A Feature-level Bias Evaluation Framework for Facial Expression Recognition Models

**Supplementary Material Overview**

1) **Section I**: Description of the data cleaning process.
2) **Section II**: Training details and performance of the deep learning models used in this study.
3) **Section III**: Detailed values of $\widetilde{V}_k^{e,\mathcal{M}}$ and all corresponding $p$-values across facial expressions and sensitive attributes for each method in our main experiments based on Swin-B.
4) **Section IV**: Experimental results for the additional network architectures.

## I. Data Cleaning

Both the facial expression and facial attribute datasets inherently involve a certain level of subjectivity and uncertainty in their annotations. As a result, ambiguous annotations may exist in the datasets, which could potentially impact our bias analysis. Manually inspecting all the data across the four datasets would be a highly time-consuming and impractical task. To overcome this challenge, we proposed to use automated filtering algorithms as an initial step.

We employed two filtering algorithms to first identify potentially ambiguous samples from the clusters in the feature space of FER and facial attribute models. These automatically flagged samples were then subjected to manual inspection. Ambiguous samples were defined as those that either: (1) are significantly distant from the centroids of their respective clusters (outliers), or (2) are located too close to samples from other clusters (overlaps). As detailed in Algorithms 1 and 2, we utilized two filtering methods: Centroid-based filtering and Closeness-based filtering. These algorithms effectively narrow down the potentially ambiguous samples for further manual examination.

After manually reviewing the samples flagged by the filtering methods, we discarded those determined to be potentially misannotated[1]. A portion of these ambiguous samples is shown in Fig. 1. For reproducibility, all discarded samples are located in the ./MisAnnotation/ directory within suppl.7z.[2]

## II. Training Details and Performance

In this section, we provide details on the hyperparameters and performance of the deep learning models used in our study. This includes eight FER models with different architectures that we have trained and evaluated on the AffectNet dataset. The architectures include ResNet-18, ResNet-34, ResNet-101, ResNet-152, Swin Transformer models (Tiny, Small, and Base), and Vision Transformer Base. All models are trained using the same set of hyperparameters. We also include six facial attribute classifiers: three trained on the FairFace dataset for gender, age, and race, and three trained on the UTKFace dataset for the same attributes. All attribute classifiers are based on ResNet-34 to maintain consistency with prior fairness studies that use these models to generate pseudo-demographic labels. These classifiers are included solely for the purpose of reproducing such previous studies. The hyperparameters used for training are listed in Table I. The performance of the FER models is reported in Table II, and the performance of the facial attribute classifiers on the AffectNet test set is shown in Table III. The code is also make available at https://github.com/Supltz/FairFER.

## III. $\widetilde{V}_k^{e,\mathcal{M}}$ and Corresponding $p$-values

We provide $\widetilde{V}_k^{e,\mathcal{M}}$ and $\widetilde{V}_k^{e,GT}$ along with their corresponding $p$-values across gender, race, and age in Tables IV, V, and VI. Table IV corresponds to Tables I and II in the main paper, Table V corresponds to Tables III and IV, and Table VI corresponds to Tables V and VI. These tables serve as a supplement to the results presented in the main paper across facial expressions and sensitive attributes for each method in our main experiments based on Swin-B.

## IV. Additional Network Architectures

All detailed results for the other network architectures are included in the JSON files located in the ./other_networks/ directory within suppl.7z. Please refer to readme.txt for more information.

---

[1] Our manual review process is also subjective, and the potentially misannotated samples represent less than 1% of the datasets. Hence, we opted to discard these samples rather than re-annotating them.

[2] We do not claim to have identified all misannotated data, as our review was limited to the samples returned by the automated filtering algorithms based on specific assumptions about ambiguous samples. We provide the filenames rather than original images to respect the copyright of the dataset providers.

**Algorithm 1** Centroids-based filtering

1: **Input:** Set of data points $X = \{x_1, x_2, \ldots, x_n\}$, corresponding labels $L = \{l_1, l_2, \ldots, l_n\}$, scale factor $s$
2: **Output:** Filtered data points $X_{filtered}$
3: **function** FILTERDATA($X, L, s$)
4:     $X_{filtered} \leftarrow \{\}$
5:     **for** each unique label $l$ in $L$ **do**
6:         $X_l \leftarrow \{x_i : l_i = l\}$
7:         $N_l \leftarrow |X_l|$
8:         $C_l \leftarrow \frac{1}{N_l} \sum_{x_i \in X_l} x_i$
9:         $\sigma_l \leftarrow \sqrt{\frac{1}{N_l} \sum_{x_i in X_l} \|x_i - C_l\|^2}$
10:       $D_l \leftarrow s \times \sigma_l$
11:       **for** each $x_i$ in $X_l$ **do**
12:           **if** $\|x_i - C_l\| \geq D_l$ **then**
13:               $X_{filtered} \leftarrow X_{filtered} \cup \{x_i\}$
14:           **end if**
15:       **end for**
16:     **end for**
17:     **return** $X_{filtered}$
18: **end function**

**Algorithm 2** Closeness-based filtering

1: **Input:** Set of data points $X = \{x_1, x_2, \ldots, x_n\}$, labels $L = \{l_1, l_2, \ldots, l_n\}$, closeness threshold $t$, probability $p$
2: **Output:** Filtered data points $X_{filtered}$
3: **function** FILTERDATA($X, L, t, p$)
4:     $X_{filtered} \leftarrow \{\}$
5:     **for** each pair $(x_i, x_j)$ in $X$ **do**
6:         **if** $\|x_i - x_j\| < t$ and $l_i \neq l_j$ **then**
7:             $N_i \leftarrow$ count of $l_i$ in $L$
8:             $N_j \leftarrow$ count of $l_j$ in $L$
9:             $r \sim U(0, 1)$
10:           **if** $r < p$ **then**
11:               **if** $N_i > N_j$ **then**
12:                   $X_{filtered} \leftarrow X_{filtered} \cup \{x_i\}$
13:               **else**
14:                   $X_{filtered} \leftarrow X_{filtered} \cup \{x_j\}$
15:               **end if**
16:           **end if**
17:         **end if**
18:     **end for**
19:     **return** $X_{filtered}$
20: **end function**

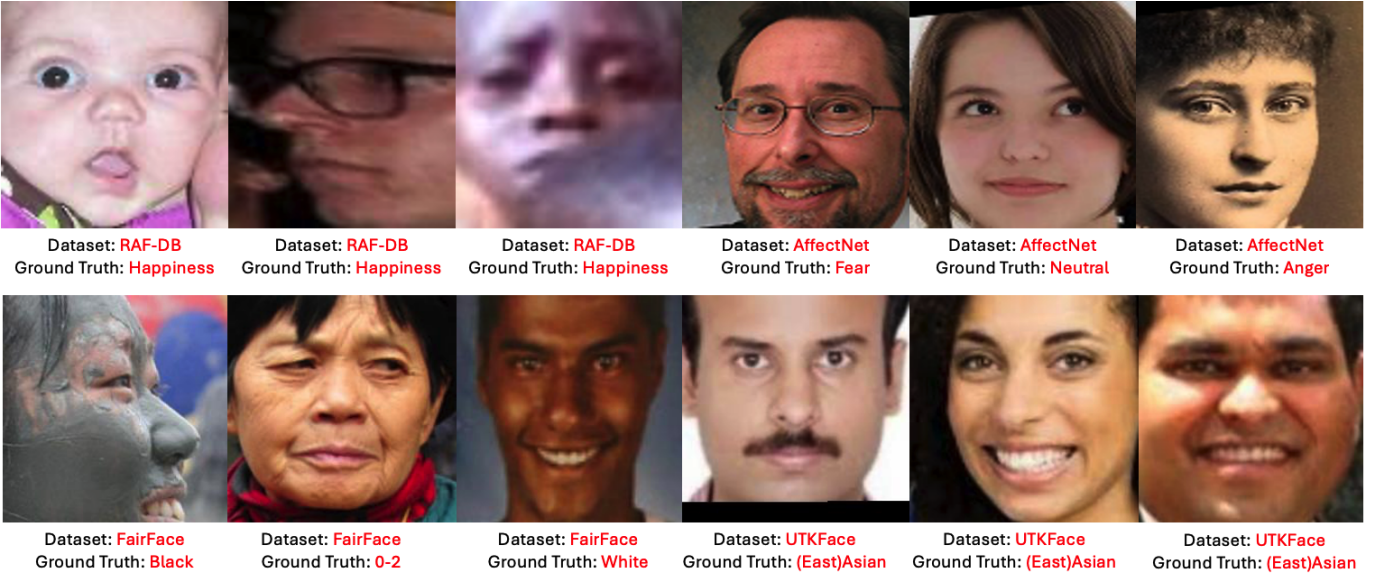

Fig. 1. Selected potentially misannotated samples in the four datasets.

TABLE I
SUMMARY OF TRAINING HYPERPARAMETERS, INCLUDING BATCH SIZE, LEARNING RATE, MOMENTUM, WEIGHT DECAY, AND EARLY STOPPING PATIENCE.

| Models \ Hyperparameters | Batch Size | Learning Rate | Momentum | Weight Decay | Patience |
|---|---|---|---|---|---|
| FER Models (AffectNet) | 64 | 2e-4 | 0.8 | 5e-3 | 5 |
| Age Model (FairFace) | 64 | 1e-4 | 0.9 | 5e-5 | 5 |
| Age Model (UTKFace) | 8 | 1e-4 | 0.9 | 5e-4 | 5 |
| Race Model (FairFace) | 64 | 1e-4 | 0.9 | 5e-5 | 5 |
| Race Model (UTKFace) | 8 | 1e-4 | 0.9 | 5e-4 | 5 |
| Gender Model (FairFace) | 64 | 1e-4 | 0.9 | 5e-4 | 5 |
| Gender Model (UTKFace) | 8 | 1e-4 | 0.9 | 5e-4 | 5 |

TABLE II
ACCURACY OF THE EIGHT FER MODELS ON THE AFFECTNET TEST SET.

| Backbone | Surprise | Fear | Disgust | Happiness | Sadness | Anger | Neutral | Avg. |
|---|---|---|---|---|---|---|---|---|
| ResNet18 | 0.798 | 0.928 | 0.504 | 0.433 | 0.406 | 0.264 | 0.606 | 0.563 |
| ResNet34 | 0.726 | 0.960 | 0.560 | 0.397 | 0.446 | 0.338 | 0.516 | 0.563 |
| ResNet101 | 0.806 | 0.946 | 0.504 | 0.357 | 0.444 | 0.334 | 0.558 | 0.564 |
| Swin-T | 0.754 | 0.952 | 0.572 | 0.419 | 0.380 | 0.278 | 0.606 | 0.566 |
| ResNet152 | 0.738 | 0.944 | 0.536 | 0.421 | 0.438 | 0.314 | 0.592 | 0.569 |
| Swin-S | 0.796 | 0.944 | 0.560 | 0.359 | 0.470 | 0.338 | 0.542 | 0.573 |
| ResNet34 | 0.802 | 0.938 | 0.534 | 0.415 | 0.494 | 0.280 | 0.548 | 0.573 |
| Swin-B | 0.762 | 0.922 | 0.580 | 0.439 | 0.424 | 0.484 | 0.638 | 0.614 |

TABLE III
ACCURACY OF THE FACIAL ATTRIBUTE MODELS ON THE AFFECTNET TEST SET.

| Models | Gender | | Race | | | | Age | | | | |
|---|---|---|---|---|---|---|---|---|---|---|---|
| | Male | Female | White | Asian | Indian | Black | 0–3 | 4–19 | 20–39 | 40–69 | 70+ |
| UTKFace | 0.834 | 0.945 | 0.851 | 0.946 | 0.673 | 0.559 | 0.590 | 0.599 | 0.789 | 0.546 | 0.926 |
| FairFace | 0.906 | 0.916 | 0.890 | 0.957 | 0.809 | 0.596 | 0.745 | 0.752 | 0.864 | 0.507 | 0.759 |

TABLE IV
THE ORIGINAL RESULTS AND THE CORRESPONDING $p$-VALUES OF BIAS EVALUATION METHODS ACROSS SEVEN FACIAL EXPRESSIONS FOR THE SENSITIVE ATTRIBUTE GENDER. THE VALUES OF $DEO^e_{(max,k)}$ AND $DiA^e_{(max,k)}$ FOR THE UNDERPERFORMING OR LESS ASSOCIATED ATTRIBUTE GROUP ARE SHOWN AS PERCENTAGES. IF THERE IS NO STATISTICALLY SIGNIFICANT BIAS BETWEEN THE REFERENCE GROUP AND THE REMAINING GROUP, THE RELATIONSHIP IS DENOTED AS {REFERENCE GROUP}/{REMAINING GROUP(S)}. RED HIGHLIGHTS INDICATE THE CLOSEST MATCH TO THE GROUND TRUTH VALUE BASED ON THE L1 DISTANCE FOR EACH EXPRESSION. (FEMALE: F, MALE: M)

| Methods \ Expressions | Ground Truth | UTKFace (pseudo) | FairFace (pseudo) | UTFFace (ours) | FairFace (ours) |
|---|---|---|---|---|---|
| | $DEO^e_{(max,k)}$ (%) / $p$-value | | | $DiA^e_{(max,k)}$ (%) / $p$-value | |
| Surprise | M/F | M/F | M/F | M/F | M/F |
| | 0/0.1234 | 0/0.0897 | 0/0.1523 | 0/0.1762 | 0/0.0956 |
| Fear | M/F | M/F | M/F | M/F | M/F |
| | 0/0.1453 | 0/0.1834 | 0/0.0678 | 0/0.1245 | 0/0.1567 |
| Disgust | Male | Male | Male | Male | **Male** |
| | 9.01/0.0234 | 7.58/0.0156 | 10.25/0.0378 | 8.59/0.0412 | **9.24/0.0289** |
| Happiness | Male | Male | Male | **Male** | Male |
| | 4.94/0.0167 | 3.03/0.0445 | 3.41/0.0323 | **5.07/0.0189** | 4.72/0.0256 |
| Sadness | Male | Male | **Male** | Male | Male |
| | 3.22/0.0312 | 1.98/0.0278 | **3.63/0.0145** | 3.54/0.0467 | 2.98/0.0234 |
| Anger | Female | Female | Female | Female | **Female** |
| | 9.30/0.0356 | 7.78/0.0123 | 8.48/0.0445 | 9.63/0.0267 | **8.99/0.0389** |
| Neutral | Male | Male | Male | **Male** | Male |
| | 2.41/0.0156 | 0.20/0.0423 | 4.33/0.0278 | **2.29/0.0345** | 2.55/0.0189 |

TABLE V
THE ORIGINAL RESULTS AND THE CORRESPONDING $p$-VALUES OF BIAS EVALUATION METHODS ACROSS SEVEN FACIAL EXPRESSIONS FOR THE SENSITIVE ATTRIBUTE RACE. THE VALUES OF $DEO^e_{(max,k)}$ AND $DiA^e_{(max,k)}$ FOR THE UNDERPERFORMING OR LESS ASSOCIATED ATTRIBUTE GROUPS ARE SHOWN AS PERCENTAGES. IF THERE IS NO STATISTICALLY SIGNIFICANT BIAS BETWEEN THE REFERENCE GROUP AND THE REMAINING GROUP, THE RELATIONSHIP IS DENOTED AS {REFERENCE GROUP}/{REMAINING GROUP(S)}. GRAY CELLS HIGHLIGHT RESULTS WHERE A METHOD IDENTIFIES THE INCORRECT REFERENCE GROUP. RED HIGHLIGHTS INDICATE THE CLOSEST MATCH TO THE GROUND TRUTH VALUE BASED ON THE AVERAGE L1 DISTANCE FOR EACH EXPRESSION. (WHITE: W, BLACK: B, ASIAN: A, INDIAN: I)

| Methods / Expressions | Ground Truth | | | UTKFace (pseudo) | | | FairFace (pseudo) | | | UTKFace (ours) | | | FairFace (ours) | | |
|---|---|---|---|---|---|---|---|---|---|---|---|---|---|---|---|
| | | | | $DEO^e_{(max,k)}$ (%) ($p$-value) | | | | | | $DiA^e_{(max,k)}$ (%) ($p$-value) | | | | | |
| Surprise | B | A | I | W | A | I | B | A | I | B | A | I | B | A | I |
| | 8.23 | 10.92 | 9.63 | 12.54 | 14.23 | 13.81 | 13.18 | 15.87 | 14.58 | 8.94 | 11.63 | 10.34 | 9.12 | 11.81 | 10.52 |
| | (0.0234) | (0.0412) | (0.0378) | (0.0156) | (0.0223) | (0.0381) | (0.0423) | (0.0167) | (0.0268) | (0.0345) | (0.0434) | (0.0123) | (0.0256) | (0.0367) | (0.0445) |
| Fear | W | B | A | W | A | I | W | B | I | W | B | A | W | B | A |
| | 7.24 | 5.62 | 16.63 | 3.84 | 8.92 | 6.73 | 4.56 | 7.34 | 5.67 | 8.13 | 6.51 | 17.52 | 7.36 | 5.74 | 16.75 |
| | (0.0167) | (0.0445) | (0.0323) | (0.0189) | (0.0256) | (0.0389) | (0.0234) | (0.0145) | (0.0467) | (0.0312) | (0.0278) | (0.0145) | (0.0289) | (0.0189) | (0.0256) |
| Disgust | W/B | W/A | I | W/B | W/A | I | W/B | W/A | I | W/B | W/A | I | W/B | W/A | I |
| | 0 | 0 | 5.72 | 0 | 0 | 8.95 | 0 | 0 | 8.89 | 0 | 0 | 6.12 | 0 | 0 | 5.98 |
| | (0.1234) | (0.0897) | (0.0423) | (0.1762) | (0.0956) | (0.0156) | (0.1523) | (0.1834) | (0.0378) | (0.0678) | (0.1245) | (0.0412) | (0.1567) | (0.0956) | (0.0289) |
| Happiness | W/B | W/A | W/I | W/B | W/A | W/I | W/B | W/A | W/I | W/B | W/A | W/I | W/B | W/A | W/I |
| | 0 | 0 | 0 | 0 | 0 | 0 | 0 | 0 | 0 | 0 | 0 | 0 | 0 | 0 | 0 |
| | (0.1523) | (0.1834) | (0.0678) | (0.1245) | (0.1567) | (0.0956) | (0.0897) | (0.1762) | (0.1523) | (0.1834) | (0.0678) | (0.1245) | (0.1567) | (0.0956) | (0.1523) |
| Sadness | W | B | I | W | A | I | B | I | A | W | B | I | W | B | I |
| | 8.23 | 10.93 | 11.82 | 6.45 | 8.23 | 4.67 | 5.89 | 3.45 | 7.82 | 8.79 | 11.49 | 12.38 | 9.07 | 11.77 | 12.66 |
| | (0.0356) | (0.0123) | (0.0445) | (0.0267) | (0.0389) | (0.0156) | (0.0423) | (0.0278) | (0.0145) | (0.0467) | (0.0234) | (0.0189) | (0.0256) | (0.0289) | (0.0389) |
| Anger | B | W/A | W/I | B | W/A | W/I | B | W/A | W/I | B | W/A | W/I | B | W/A | W/I |
| | 7.81 | 0 | 0 | 2.10 | 0 | 0 | 1.82 | 0 | 0 | 8.25 | 0 | 0 | 7.89 | 0 | 0 |
| | (0.0289) | (0.1523) | (0.0897) | (0.0412) | (0.1762) | (0.0956) | (0.0378) | (0.1523) | (0.1834) | (0.0156) | (0.0678) | (0.1245) | (0.0223) | (0.1567) | (0.0956) |
| Neutral | W | B/A | I | W | B/A | I | W | B/A | I | W | B/A | I | W | B/A | I |
| | 4.23 | 0 | 5.82 | 9.60 | 0 | 0.45 | 8.34 | 0 | 1.71 | 4.70 | 0 | 6.29 | 4.38 | 0 | 5.97 |
| | (0.0345) | (0.1834) | (0.0167) | (0.0434) | (0.0678) | (0.0445) | (0.0123) | (0.1245) | (0.0356) | (0.0267) | (0.1567) | (0.0389) | (0.0156) | (0.0956) | (0.0423) |

TABLE VI
THE ORIGINAL RESULTS AND THE CORRESPONDING $p$-VALUES OF BIAS EVALUATION METHODS ACROSS SEVEN FACIAL EXPRESSIONS FOR THE SENSITIVE ATTRIBUTE AGE. THE VALUES OF $DEO^e_{(max,k)}$ AND $DiA^e_{(max,k)}$ FOR THE UNDERPERFORMING OR LESS ASSOCIATED ATTRIBUTE GROUPS ARE SHOWN AS PERCENTAGES. IF THERE IS NO STATISTICALLY SIGNIFICANT BIAS BETWEEN THE REFERENCE AND REMAINING GROUP(S), THE RELATIONSHIP IS DENOTED AS {REFERENCE GROUP}/{REMAINING GROUP(S)}. GRAY CELLS INDICATE CASES WHERE A METHOD IDENTIFIES THE INCORRECT REFERENCE GROUP. RED HIGHLIGHTS INDICATE THE CLOSEST MATCH TO THE GROUND TRUTH VALUE BASED ON THE AVERAGE L1 DISTANCE FOR EACH EXPRESSION. ($A_1$: 0–3, $A_2$: 4–19, $A_3$: 20–39, $A_4$: 40–69, $A_5$: 70+ YEARS)

| Expressions \ Methods | Ground Truth | | | | UTKFace (pseudo) | | | | FairFace (pseudo) | | | | UTFFace (ours) | | | | FairFace (ours) | | | |
|---|---|---|---|---|---|---|---|---|---|---|---|---|---|---|---|---|---|---|---|---|
| | $DEO^e_{(max,k)}$ (%) ($p$-value) | | | | | | | | | | | | $DiA^e_{(max,k)}$ (%) ($p$-value) | | | | | | | |
| Surprise | $A_1/A_5$ | $A_2$ | $A_3$ | $A_4$ | $A_1/A_2$ | $A_3$ | $A_4$ | $A_5$ | $A_1/A_3$ | $A_2$ | $A_4$ | $A_5$ | $A_1/A_5$ | $A_2$ | $A_3$ | $A_4$ | $A_1/A_5$ | $A_2$ | $A_3$ | $A_4$ |
| | 0 | 13.24 | 15.31 | 14.23 | 0 | 19.24 | 10.31 | 17.23 | 0 | 15.74 | 12.31 | 16.23 | 0 | 13.55 | 15.61 | 14.54 | 0 | 13.68 | 15.75 | 14.67 |
| | (0.1234) | (0.0345) | (0.0123) | (0.0234) | (0.0876) | (0.0445) | (0.0156) | (0.0389) | (0.1523) | (0.0267) | (0.0312) | (0.0423) | (0.1762) | (0.0189) | (0.0234) | (0.0445) | (0.0956) | (0.0156) | (0.0378) | (0.0412) |
| Fear | $A_2$ | $A_3$ | $A_4$ | $A_5$ | $A_1$ | $A_2/A_3$ | $A_4$ | $A_5$ | $A_2$ | $A_3$ | $A_4$ | $A_5$ | $A_2$ | $A_3$ | $A_4$ | $A_5$ | $A_2$ | $A_3$ | $A_4$ | $A_5$ |
| | 13.34 | 14.13 | 14.42 | 25.31 | 19.13 | 0 | 8.42 | 31.31 | 18.46 | 19.25 | 19.54 | 30.43 | 13.69 | 14.48 | 14.77 | 25.66 | 13.76 | 14.55 | 14.84 | 25.73 |
| | (0.0289) | (0.0345) | (0.0156) | (0.0423) | (0.0189) | (0.1245) | (0.0467) | (0.0234) | (0.0189) | (0.0256) | (0.0389) | (0.0156) | (0.0423) | (0.0278) | (0.0145) | (0.0467) | (0.0234) | (0.0189) | (0.0256) | (0.0389) |
| Disgust | $A_1$ | $A_2/A_3$ | $A_4$ | $A_5$ | $A_3$ | $A_1/A_2$ | $A_4$ | $A_5$ | $A_1$ | $A_2/A_3$ | $A_4$ | $A_5$ | $A_1$ | $A_2/A_3$ | $A_4$ | $A_5$ | $A_1$ | $A_2/A_3$ | $A_4$ | $A_5$ |
| | 14.32 | 0 | 17.14 | 34.23 | 20.32 | 0 | 23.14 | 28.23 | 20.15 | 0 | 22.97 | 40.06 | 14.90 | 0 | 17.72 | 34.81 | 14.73 | 0 | 17.55 | 34.64 |
| | (0.0356) | (0.1523) | (0.0123) | (0.0445) | (0.0267) | (0.0897) | (0.0389) | (0.0156) | (0.0423) | (0.1834) | (0.0278) | (0.0145) | (0.0467) | (0.0678) | (0.0234) | (0.0189) | (0.0256) | (0.1567) | (0.0389) | (0.0156) |
| Happiness | $A_1/A_2$ | $A_2/A_3$ | $A_4$ | $A_2/A_5$ | $A_1/A_2$ | $A_2/A_3$ | $A_4$ | $A_2/A_5$ | $A_1/A_2$ | $A_2/A_3$ | $A_4$ | $A_2/A_5$ | $A_1/A_2$ | $A_2/A_3$ | $A_4$ | $A_2/A_5$ | $A_1/A_2$ | $A_2/A_3$ | $A_4$ | $A_2/A_5$ |
| | 0 | 0 | 3.14 | 0 | 0 | 0 | 11.14 | 0 | 0 | 0 | 8.64 | 0 | 0 | 0 | 3.44 | 0 | 0 | 0 | 3.63 | 0 |
| | (0.0956) | (0.1523) | (0.0445) | (0.1834) | (0.0678) | (0.1245) | (0.0467) | (0.1567) | (0.0956) | (0.1523) | (0.0389) | (0.1834) | (0.0678) | (0.1245) | (0.0467) | (0.1567) | (0.0956) | (0.1523) | (0.0389) | (0.1834) |
| Sadness | $A_2$ | $A_3$ | $A_4$ | $A_5$ | $A_1$ | $A_2$ | $A_3$ | $A_5$ | $A_1$ | $A_2$ | $A_4$ | $A_5$ | $A_2$ | $A_3$ | $A_4$ | $A_5$ | $A_2$ | $A_3$ | $A_4$ | $A_5$ |
| | 17.23 | 33.14 | 31.42 | 28.13 | 22.23 | 28.14 | 35.42 | 24.13 | 14.73 | 35.64 | 28.92 | 30.63 | 17.75 | 33.66 | 31.94 | 28.65 | 17.68 | 33.59 | 31.87 | 28.58 |
| | (0.0234) | (0.0412) | (0.0378) | (0.0156) | (0.0223) | (0.0381) | (0.0423) | (0.0167) | (0.0268) | (0.0123) | (0.0434) | (0.0367) | (0.0445) | (0.0256) | (0.0289) | (0.0389) | (0.0256) | (0.0367) | (0.0445) | (0.0123) |
| Anger | $A_1/A_2$ | $A_3$ | $A_4$ | $A_5$ | $A_1$ | $A_2$ | $A_4$ | $A_5$ | $A_1$ | $A_3$ | $A_4$ | $A_5$ | $A_1/A_2$ | $A_3$ | $A_4$ | $A_5$ | $A_1/A_2$ | $A_3$ | $A_4$ | $A_5$ |
| | 0 | 21.31 | 14.23 | 30.14 | 18.24 | 16.31 | 19.23 | 35.14 | 15.74 | 23.81 | 11.73 | 32.64 | 0 | 21.82 | 14.74 | 30.65 | 0 | 21.69 | 14.61 | 30.52 |
| | (0.1523) | (0.0467) | (0.0234) | (0.0189) | (0.0256) | (0.0389) | (0.0156) | (0.0423) | (0.0278) | (0.0145) | (0.0467) | (0.0234) | (0.1834) | (0.0189) | (0.0256) | (0.0389) | (0.0678) | (0.0156) | (0.0423) | (0.0278) |
| Neutral | $A_1$ | $A_2/A_3$ | $A_4$ | $A_5$ | $A_1$ | $A_2$ | $A_4$ | $A_5$ | $A_1$ | $A_2/A_3$ | $A_4$ | $A_5$ | $A_1$ | $A_2/A_3$ | $A_4$ | $A_5$ | $A_1$ | $A_2/A_3$ | $A_4$ | $A_5$ |
| | 13.42 | 0 | 14.31 | 25.24 | 22.25 | 0 | 23.14 | 34.07 | 19.36 | 0 | 20.25 | 31.18 | 13.88 | 0 | 14.77 | 25.70 | 13.79 | 0 | 14.68 | 25.61 |
| | (0.0145) | (0.1467) | (0.0234) | (0.0189) | (0.0256) | (0.0389) | (0.0156) | (0.0423) | (0.0278) | (0.1245) | (0.0467) | (0.0234) | (0.0189) | (0.1256) | (0.0389) | (0.0156) | (0.0423) | (0.1678) | (0.0278) | (0.0145) |



\end{document}